\documentclass[runningheads]{llncs}

\usepackage{eccv}

\usepackage{eccvabbrv}

\usepackage{graphicx}
\usepackage{booktabs}
\usepackage{tcolorbox}
\tcbuselibrary{skins}

\usepackage[accsupp]{axessibility}  % Improves PDF readability for those with disabilities.

\usepackage[breaklinks,colorlinks,citecolor=eccvblue]{hyperref}
\usepackage{microtype}
\usepackage{geometry}
\newgeometry{left=1.5in,right=1.5in,top=1in,bottom=1in}
\usepackage{orcidlink}

\usepackage{hyperref}
\usepackage{titletoc}

\usepackage{amsmath,amsfonts,bm}

\def\eqref#1{equation~\ref{#1}}
\def\1{\bm{1}}

\DeclareMathAlphabet{\mathsfit}{\encodingdefault}{\sfdefault}{m}{sl}
\SetMathAlphabet{\mathsfit}{bold}{\encodingdefault}{\sfdefault}{bx}{n}

\usepackage{url}
\usepackage{tikz}
\usepackage{booktabs}
\usepackage{ulem}
\usepackage{multirow}
\usepackage[table,xcdraw]{xcolor}

\definecolor{lightorange}{RGB}{250, 237, 205}
\definecolor{lightblue}{RGB}{202, 240, 248}
\definecolor{lightgreen}{RGB}{233, 237, 201}

\newcommand{\finding}[2]{
    \vspace{-0.1cm}
    \begin{tcolorbox}[
        colback=white!90!gray,     % Soft ivory background
        colframe=teal!60!black,     % Deep navy frame color
        arc=5pt,                    % Less rounded corners for a modern look
        boxsep=5pt,                 % Inner margin between text and frame
        left=10pt,                  % Left margin within the box
        right=10pt,                 % Right margin within the box
        top=2pt,                    % Top margin within the box
        bottom=2pt,                 % Bottom margin within the box
        boxrule=0.8pt,              % Frame thickness
        enhanced jigsaw             % Allows for the shadow effect and better arcs
    ]
    \vspace{-0.1cm}
        \paragraph{\textbf{\textit{Finding #1:}}} #2
    \vspace{-0.1cm}
    \end{tcolorbox}
    \vspace{-0.1cm}
}

\usepackage{listings}
\usepackage{enumitem}
\tcbuselibrary{breakable}

\newtcolorbox{promptbox}[1][]{
  colback=gray!5,
  colframe=gray!50,
  boxrule=0.8pt,
  arc=4pt,
  left=8pt, right=8pt, top=6pt, bottom=6pt,
  fonttitle=\bfseries\small,
  breakable,  % This will now work!
  #1  
}

\definecolor{myblue}{HTML}{FDF5E0}
\definecolor{mygray}{HTML}{DBE2E9}
\definecolor{mygreen}{HTML}{E6F3FC}

\usepackage[capitalize]{cleveref}
\crefname{section}{Sec.}{Secs.}
\Crefname{section}{Section}{Sections}
\Crefname{table}{Table}{Tables}
\crefname{table}{Tab.}{Tabs.}
\usepackage{makecell}
\usepackage{amsmath}
\usepackage{amssymb}
\usepackage{mathtools}
\usepackage[textsize=tiny]{todonotes}

\begin{document}

% ---------------------------------------------------------------
\title{Tinted Frames: Question Framing Blinds\\Vision-Language Models} 

\titlerunning{Tinted Frames: Question Framing Blinds Vision-Language Models}

\author{Wan-Cyuan Fan\inst{1,3}, Jiayun Luo$^\dagger$\inst{1,3}, Declan Kutscher$^\dagger$\inst{2},\\Leonid Sigal\inst{1,3}, Ritwik Gupta\inst{2}}

{\def\thefootnote{}\footnotetext{$^\dagger$Equal contribution. Corresponding author \texttt{wancyuan@cs.ubc.ca}.\\$^\star$\href{https://davidhalladay.github.io/tinted_frames_demo}{Project Page}}}

\authorrunning{Fan, Luo, Kutscher, et al.}

\institute{University of British Columbia  \and
University of California, Berkeley \and
Vector Institute for AI}

\maketitle

% Color code: green: #4DBFB6, #73BC81, blue: #8B9FCD, yellow: #EDB03B, #ECAB2D, #F8D19C, red: #FF6F66, purple, #7470AF

\begin{abstract}
Vision-Language Models (VLMs) have been shown to be blind, often underutilizing their visual inputs even on tasks that require visual reasoning. In this work, we demonstrate that VLMs are \textit{selectively} blind. They modulate the amount of attention applied to visual inputs based on linguistic framing even when alternative framings demand identical visual reasoning.
Using visual attention as a probe, we quantify how framing alters both the amount and distribution of attention over the image.
Constrained framings, such as multiple choice and yes/no, induce substantially lower attention to image context compared to open-ended, reduce focus on task-relevant regions, and shift attention towards uninformative tokens.
We further demonstrate that this attention misallocation is the principal cause of degraded accuracy and cross-framing inconsistency. 
Building on this mechanistic insight, we introduce a lightweight prompt-tuning method using learnable tokens that encourages the robust, visually grounded attention patterns observed in open-ended settings, improving visual grounding and improving performance across framings. 

\end{abstract}

\section{Introduction}

Complex, multi-modal reasoning tasks have been the driving force behind contemporary vision-language model (VLM) development. As these models tackle increasingly difficult real-world datasets, it is critical that their reasoning and responses are appropriately grounded in visual evidence. However, despite their impressive performance on simple benchmarks, recent research has revealed that the visual capability of these systems is largely a function of text priors and biases. VLMs are ``blind'' and exhibit distinct failures in visual grounding, raising fundamental questions about whether they are reasoning over the image or merely leveraging powerful language priors to generate plausible answers.

Recent work characterizes these issues as a problem of visual disengagement and structural bias. Studies~\cite{tong2024cambrian, rahmanzadehgervi2024vision} have shown that VLMs assign little attention to visual tokens, generating responses driven primarily by textual context rather than visual evidence. This lack of attention is not uniformly distributed. Models frequently allocate disproportionately high attention weights to visual attention sink tokens~\cite{kang2025see, luo2025sink, kaduri2025s}, semantically meaningless background tokens, while also exhibiting severe spatial biases. For instance, artifacts from positional encodings (i.e., RoPE~\cite{su2023enhanced}) and causal attention mechanisms can create effective ``blind spots''~\cite{zhu2025bias, tian2025identifying, wang2025circle}, resulting in the neglect of specific image regions regardless of their semantic importance. However, these analyses have typically been done holistically, averaging observations across heterogeneous benchmarks. This perspective implies that such blindness is a static, inherent flaw of the model architecture. While there is ample evidence that prompting impacts model accuracy \cite{gu2023systematic,schmalfuss2025parc,liangprompt}, there is little evidence that suggests that visual perception process itself is impacted by this, or that a simple question framing can induce such behavior.

\begin{figure*}[t]
  \centering
  \includegraphics[page=20, trim={15 30 15 30}, clip, width=0.95\textwidth]{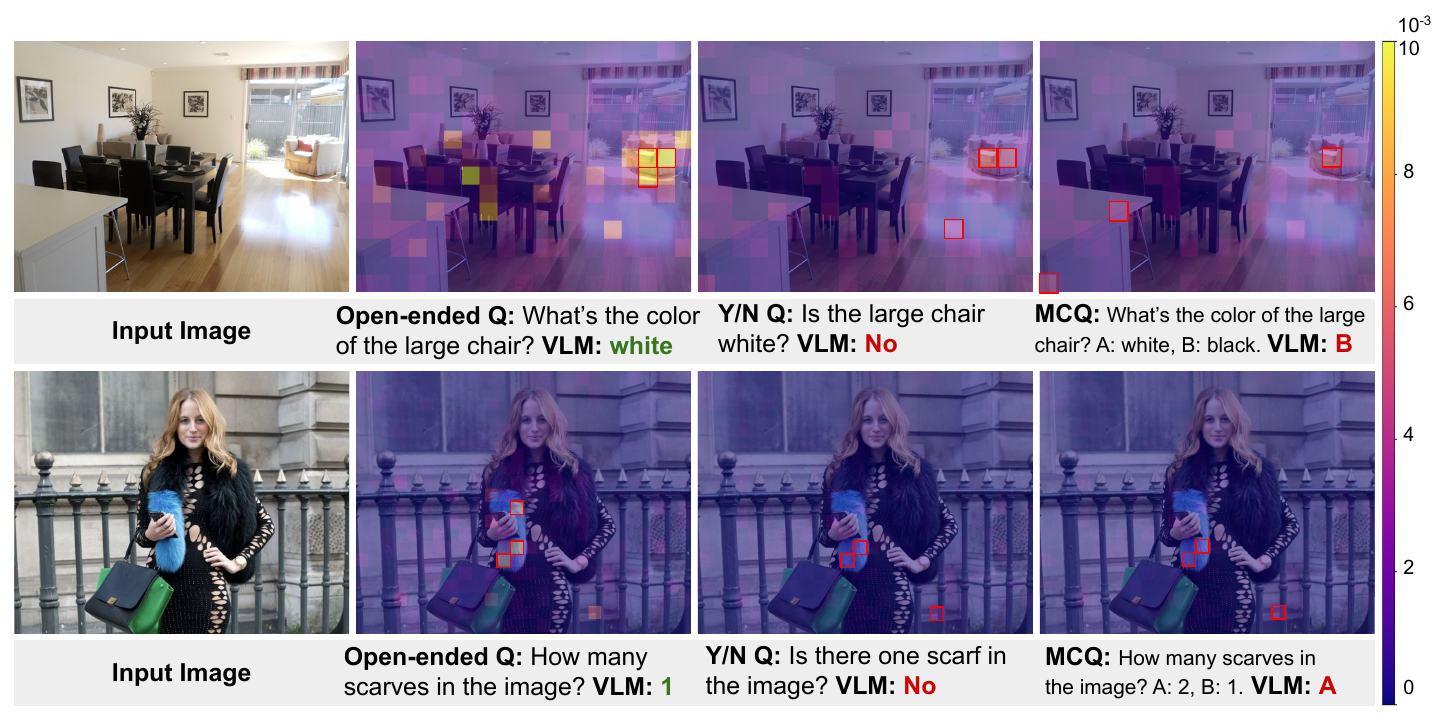}
  % \vspace{-3mm}
  \caption{\textbf{VLM grounding changes as a function of question framing.} Attention maps reveal that while the model actively attends to the target object (the chair) during open-ended generation, it exhibits disengagement and misallocation when the same question is posed as a Yes/No or MCQ task. Note that we employ attention rollout rather than averaging attention weights across layers and token. By recursively rolling out attention matrices, we trace information propagation from inputs of early layers to output embeddings. Qwen2.5-VL-7B is used for the visualization. The top 3 tokens with the highest attention are highlighted in red boxes, and the minimum and maximum values of the linear colormap are set to the same value for all images.}
  \label{fig:teasers}
  % \vspace{-5mm}
\end{figure*} 

In this work, we demonstrate that VLMs are \textit{selectively} blind. They decide how much to look at an image based on the textual framing of the question---such as open-ended, Yes/No, or multiple-choice---despite different framings requiring the same visual concepts to answer correctly. 
We study this phenomenon mechanistically, utilizing attention rollout~\cite{abnar2020quantifying} to estimate the visual information propagation and visualize the attention map where the model attends when making output decisions. We hypothesize that alternative framings impact model performance \textit{indirectly} via deviations in visual attention.

Since open-ended, Yes/No, and multiple-choice question (MCQ) are the three dominant framings used to evaluate VLMs across major benchmarks, this framing-dependent behavior has direct implications for how we assess model capabilities. 

We conduct a three-part analysis. First we establish and quantify the impact of framing on the accuracy by introducing \textit{cross-framing inconsistency} as a diagnostic measure. By posing semantically equivalent questions across all formats, we find that models which correctly answer open-ended questions frequently fail their constrained counterparts, especially for tasks involving object grounding.

Second, we find that framing impacts both amount and distribution of attention.
We find that constrained framings trigger a shift in visual attention strategy, reducing overall attention on the image and redirecting attention away from task-relevant regions.  Finally, through intervention on attention, we confirm that indeed the impact of the framing on the accuracy is induced by the shift in visual attention.

Armed with these findings, we propose a lightweight prompt-tuning mitigation strategy that learns a small set of soft tokens to realign the visual attention of constrained framings to match the robust patterns of open-ended setting. Our mitigation strategy restores visual grounding and yields consistent improvements across multiple models and benchmarks without modifying model weights.

\section{Related Work}

Vision-language models (VLMs)~\cite{li2024llava, bai2025qwen3, zeng2025glm, meta2024llama, team2025gemma, abdin2024phi, zhu2025internvl3} have rapidly advanced from simple captioning to complex multimodal reasoning, but \textit{how do we know they truly understand what they see?} 
Since contemporary VLMs output free-form language, it is difficult to disentangle visual understanding and reasoning from linguistic shortcuts~\cite{lin2023revisiting}. 
Answering this question requires looking beyond output accuracy, into whether these models truly ground their reasoning in what they see, and what factors might cause that grounding to break down.

\subsubsection{Visual Grounding in VLMs.}
Visual grounding, the ability to localize textual concepts within an image, has been long-standing goal in computer vision. Classical architectures such as object detectors~\cite{ren2015faster, carion2020end, he2017mask} are explicitly trained to produce spatial localizations. More recently, vision transformers have been shown to develop interpretable attention patterns that correlate with object boundaries~\cite{caron2021emerging}. 
VLMs embed these vision encoders but are trained end-to-end for language generation. Therefore, spatial grounding is an implicit learning task as opposed to a primary objective. Yet, recent works~\cite{kang2025see, fu2025hidden} have shown that this implicit grounding is often unreliable. VLMs can produce correct answers while attending to irrelevant regions, suggesting that strong benchmark performance does not guarantee genuine visual understanding. In this work, we establish a mechanistic link between question framing, visual attention, and output quality.

\subsubsection{Visual Disengagement and Bias in VLMs.}
A growing amount of work documents visual shortcomings of VLMs. Studies~\cite{zhang2024redundancy, tong2024cambrian} have shown that these models often allocate much lower attention to visual content than to textual ones, potentially generating responses driven by language priors rather than visual evidence. Others~\cite{kang2025see, luo2025sink} have identified that models disproportionately attend to semantically meaningless visual tokens when performing visual reasoning tasks, further diluting visual engagement on the area of interest. 
Beyond attention allocation, systematic spatial biases from rotary position embedding (RoPE), causal attention masks, and data distribution create effective blind spots~\cite{tian2025identifying}, causing models to neglect certain image regions regardless of semantic importance.  
These findings paint a picture of visual blindness as a general and static property.
In this work, we find that visual disengagement is dynamic and conditional on linguistic framing. VLMs attend to images well under open-ended framings, but do not under alternatives. Therefore, this work reframes existing findings from ``the model cannot see'' to ``the model decides not to see.''

\subsubsection{Prompt Sensitivity.}
While human-in-the-loop evaluation~\cite{chiang2024chatbot} offers a direct measure of model quality based on human preference, it does not scale to systematic probing of specific visual capabilities. Visual capabilities are benchmarked via targeted probes---partially due to convenience of evaluation. An implicit assumption in existing benchmarks is that framing is a neutral container: a model that understands the scene should answer correctly regardless of how the question is asked. But is this assumption warranted?
VLMs are known to be sensitive to how questions are phrased~\cite{chou2025mm}. Prior work has documented a range of within-format perturbations: MCQ option ordering effects~\cite{pezeshkpour2024large}, yes-bias~\cite{li2023evaluating}, negation bias~\cite{alhamoud2025vision}, and paraphrase inconsistency~ \cite{chou2025mm}. These studies vary the surface wording while keeping the question format fixed. Framing, by contrast, is a stronger structural shift: it changes the format itself (e.g., from open-ended to Yes/No or MCQ) while preserving the underlying semantic question. This axis of sensitivity has received comparatively little attention. Moreover, existing studies~\cite{chou2025mm, shah2025analyzing} primarily measure sensitivity at the output level, through accuracy drops and answer distribution shifts. This works explores the mechanism of \textit{how} framing reshapes models' visual processing flow.

\section{Hypothesis on Framing-Attention Influence}
\label{sec:hypothesis}

\begin{figure*}[t]
  \centering
  \includegraphics[page=9, trim={100 150 100 150}, clip, width=0.98\textwidth]{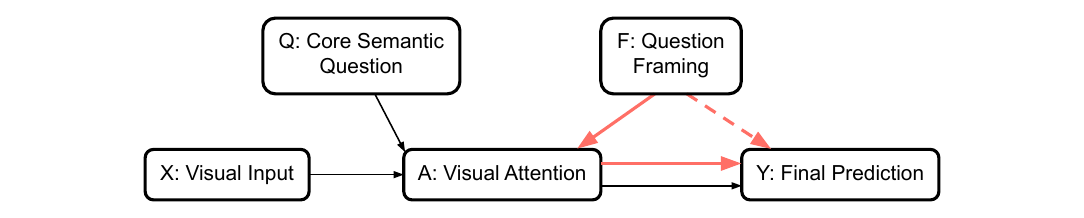}
  % \vspace{-4mm}
  \caption{{\bf Illustration of impact of question framing.} We hypothesize that question framing influences model predictions \textit{through} visual attention. Framing alters attention allocation (F$\rightarrow$A), which in turn degrades prediction quality (A$\rightarrow$Y).} %  strategy.}
  \label{fig:causal_graph}
  % \vspace{-6mm}
\end{figure*}

We illustrate an ideal processing chain for VLMs in~\cref{fig:causal_graph}. When performing visual question answering, both text and image modalities should impact visual attention, which in turn affects the final prediction. The semantics of the question are often independent of the question framing. Therefore, there should be no latent factor affecting visual attention.

However, we posit that framing directly impacts visual attention (\textbf{F$\rightarrow$A}).
Predictions follow from visual attention (\textbf{A$\rightarrow$Y}), but framing affects silently degrade the quality of this attention. Together, these form a joint pathway (\textbf{F$\rightarrow$A$\rightarrow$Y}). Therefore, a latent relationship exists between framing and the final prediction (\textbf{F$\rightarrow$Y}). For a robust VLM, none of these pathways should exist; their presence reveals that current models rely on shallow, framing-dependent heuristics rather than genuine visual understanding.

We investigate the impact and existence of each pathway in turn. \cref{sec:f2a2y} examines the overall effect of framing on predictions (\textbf{F$\rightarrow$Y}). \cref{sec:analysis} analyzes the pathway through visual attention (\textbf{F$\rightarrow$A$\rightarrow$Y}). Finally, in~\cref{sec:methodology}, we present a prompt-tuning method that realign the visual attention to restore robust predictions.

\section{Cross-Framing Inconsistency (\textbf{F$\rightarrow$Y})}
\label{sec:f2a2y}

Before examining the internal mechanisms behind framing effects, we first ask a simpler question: does question framing affect the model's final prediction (\textbf{F$\rightarrow$Y} in~\cref{fig:causal_graph})? 
To study this, we use 
% Our key insight is to use 
open-ended generation as the anchor. Among the three standard evaluation formats, open-ended questioning provides a natural anchor: without candidate options to select from, the model must generate the answer through free-form reasoning, making it less likely to succeed relying on the prior knowledge alone. If a model answers an open-ended question correctly but fails when the same question is reframed as Yes/No or MCQ, this inconsistency is unlikely to stem from a fundamental lack of visual understanding and is more likely driven by the framing itself. We formalize this as \textit{cross-framing inconsistency}: the rate at which a model fails to maintain the correct answer, as per the open-ended question, under Yes/No and MCQ framings.

\subsubsection{Evaluation Protocol.} Our protocol is illustrated in~\cref{fig:sec3_inconsistency} (left). We evaluate on GQA~\cite{hudson2019gqa}, a general VQA benchmark, and SeedBench~\cite{li2023seed}, which contains diverse visual reasoning tasks. For SeedBench, we remove the original multiple-choice options to form open-ended questions. We query the model with these open-ended questions and retain only correctly answered samples. We then use GPT-5.1 to construct semantically equivalent Yes/No reformulations from the correct answer and re-query the model, measuring whether the correct answer is preserved. Thus, the inconsistency rate is calculated based on cases where the open-ended question was answered correctly, but either the Yes/No or MCQ counterpart was incorrect. Details of the rephrasing procedure and evaluation are provided in the supplementary material.

\subsubsection{Results.}
\begin{figure}[t]
  \centering
  % --- Left Side: Figure ---
  \begin{minipage}{0.42\textwidth}
    \centering
    \includegraphics[page=4, trim={130 80 130 80}, clip, width=\linewidth]{figures/sources/main.pdf}
  \end{minipage}
  \hfill
  % % --- Right Side: Table ---
  \begin{minipage}{0.56\textwidth}
    \centering
    \includegraphics[page=1, trim={5 0 5 0}, clip, width=\linewidth]{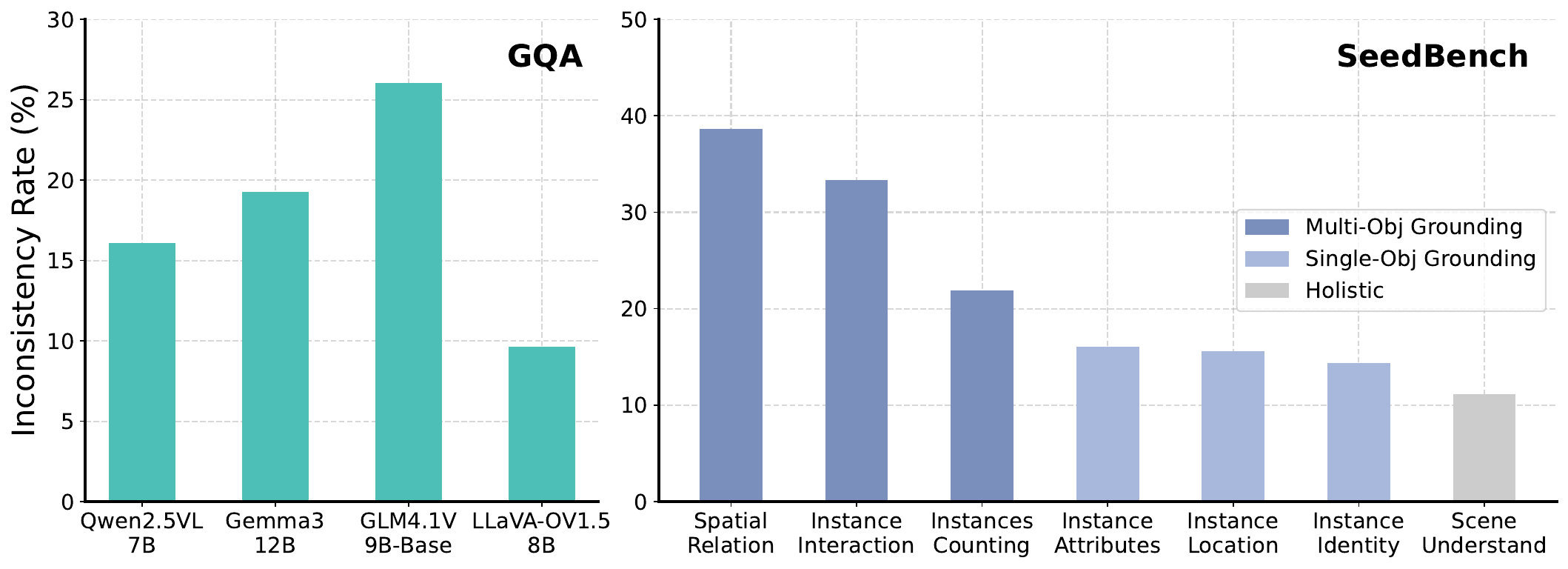}
    % \vspace{-4mm}
  \end{minipage}
  % \vspace{-4mm}
  \caption{\textbf{Tested open-source VLMs have a significant inconsistency rates across framings and task types.} (Left) Cross-framing inconsistency is evaluated by reframing questions with an LLM. (Right) With Qwen2.5-VL-7B, inconsistency is up to 26\% on GQA and 38\% on SeedBench.}
  \label{fig:sec3_inconsistency}
  % \vspace{-6mm}
\end{figure}

As shown in~\cref{fig:sec3_inconsistency} (right), the results reveal a surprising degree of inconsistency across all tested VLMs. On GQA, Qwen2.5-VL~\cite{bai2025qwen25vltechnicalreport}, Gemma3~\cite{team2025gemma}, GLM4.1V~\cite{zeng2025glm} exhibit more than $15\%$ cross-framing inconsistency, meaning the model fails to preserve its own correct answers under constrained framing for nearly one in six questions. The task-level breakdown on SeedBench using Qwen2.5-VL-7B is particularly revealing: 

\finding{1}{Tasks requiring object grounding exhibit the highest inconsistency rates, with multiple-object grounding tasks such as spatial relation and counting being the most affected, suggesting that constrained framing is most damaging where visual grounding matters most.} 

\noindent
These results confirm the existence of the connection \textbf{F$\rightarrow$Y}, establishing that framing alters predictions. 

\section{Impact of Framing on Visual Attention (\textbf{F$\rightarrow$A$\rightarrow$Y})}
\label{sec:analysis}

Having confirmed that framing alters predictions (\textbf{F$\rightarrow$Y}), we now investigate the specific internal mechanisms that contribute to this. Specifically, we ask whether and how framing reshapes the model's visual attention (\textbf{F$\rightarrow$A}), and whether any such attention shift is a primary driver of prediction failures (\textbf{A$\rightarrow$Y})?

\subsubsection{Choice of datasets and models.}
We conduct our analysis on two benchmarks selected for their spatial annotations, which enable precise mapping of attention to semantic regions. GQA~\cite{hudson2019gqa} is a general-purpose visual question answering benchmark built upon the Visual Genome dataset, providing dense semantic annotations including bounding boxes for target objects and scene graph representations that capture spatial relationships between visual entities. V$^*$~\cite{wu2024v} is a high-resolution visual grounding benchmark consisting of around 300 carefully curated samples that require fine-grained spatial reasoning in MCQ format, with bounding box annotations for target regions.

To isolate the impact of task framing from variations in question content, we employ a controlled generation approach. For each sample, we generate three distinct framing variants: open-ended, Yes/No, and MCQ, ensuring that the underlying visual reasoning required remains constant while only the output format changes. We utilize GPT-5.1 to rephrase the original samples into these target formats. For GQA, we leverage the ground-truth scene graph and object annotations to prompt the GPT, ensuring that generated Yes/No and MCQ distractors are factually consistent without requiring visual access. Detailed prompt templates and our human verification process are provided in the Supplementary Material. After filtering, we curate a final dataset of 10k unique semantic queries for GQA and the full 300 samples for V$^*$. With three framing variants per query, this yields 30k samples for GQA and 900 for V$^*$. We denote the resulting framing-controlled datasets as \textbf{GQA$^\text{F}$} and \textbf{V$^{*\text{F}}$} to distinguish them from the original benchmarks.

We primarily focus our analysis on Qwen2.5-VL 7B~\cite{bai2025qwen25vltechnicalreport}, with extended results covering Gemma3~\cite{team2025gemma} and LLaVA-OneVision1.5~\cite{an2025llava} in the supplementary.

\subsubsection{Visual attention aggregation.} 
To quantify the visual reliance of VLMs in generation, we employ attention rollout~\cite{abnar2020quantifying} rather than simple attention averaging across layers and tokens. By recursively ``rolling out'' attention matrices, we trace visual information propagation from input visual tokens to output embeddings, accounting for both direct attention and indirect pathways via residual connections. Crucially, we apply receptive field normalization to preserve causality during attention map aggregation, as required for autoregressive transformers~\cite{abnar2020quantifying}.

Formally, let $\mathbf{W}_{att}^{(\ell)} \in \mathbb{R}^{N \times N}$ be the raw attention matrix at layer $\ell$, where rows correspond to query tokens and columns to key tokens. Following \cite{abnar2020quantifying}, we account for residual connections by defining the adjusted attention matrix as $\mathbf{A}^{(\ell)} = 0.5 \mathbf{W}_{att}^{(\ell)} + 0.5 \mathbf{I}$,
where $\mathbf{I}$ denotes the identity matrix.

To address the bias arising from causal masking as discussed in previous work~\cite{wang2024eliminating}, we then apply receptive field normalization. Specifically, we scale the key tokens by $\mathbf{A}^{(\ell)}$ based on their receptive field size $\mathcal{S}$ to ensure unbiased probability mass propagation. We then re-normalize the rows to ensure the resulting matrix is row stochastic matrix before the recursive rollout:

\begin{equation}
    \mathbf{R}^{(\ell)} = \left(\mathcal{N} (\mathbf{A}^{(\ell)} \cdot \text{diag}(\mathcal{S})) \cdot \mathbf{R}^{(\ell-1)} \right),
\end{equation}

\noindent
where $\mathbf{R}^{(0)} = \mathbf{I}$. The operator $\mathcal{N}(\cdot)$ performs row-wise normalization, ensuring that each row sums to 1. The final cumulative product $\mathbf{R}^{(L)}$ represents a valid stochastic matrix capturing the effective information flow between all token pairs, where $L$ denotes total number of layers for rollout. To quantify visual reliance, we extract the specific sub-matrix connecting the generated output tokens (queries) to the input visual tokens (keys). We define the final \textbf{Visual Energy} by aggregating the probability mass within this sub-matrix; a higher total summation indicates a stronger reliance on visual content during generation.

\begin{figure}[t]
  \centering
  % Top Image
  \includegraphics[page=1, trim=5 5 5 5, clip, width=0.99\textwidth]{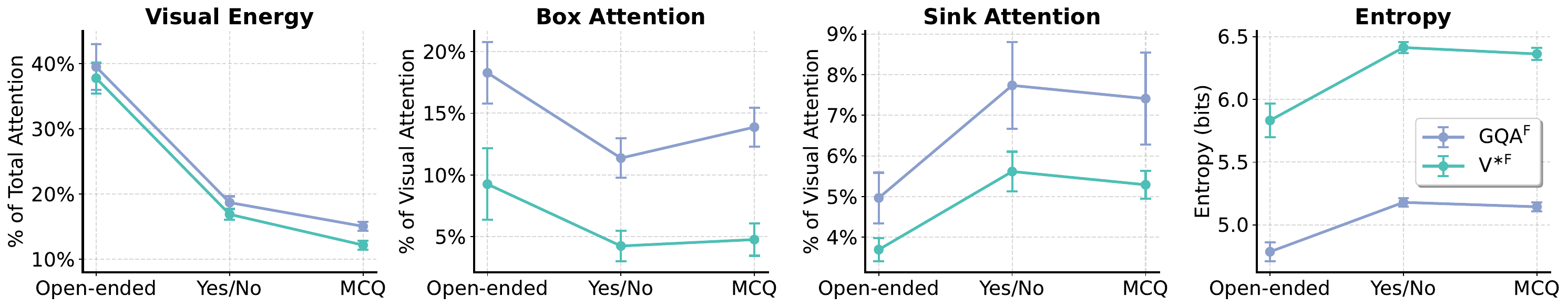}
  \\[2ex] % Adds a small vertical gap
  % Bottom Image
  \includegraphics[page=1, trim=5 5 5 5, clip, width=0.98\textwidth]{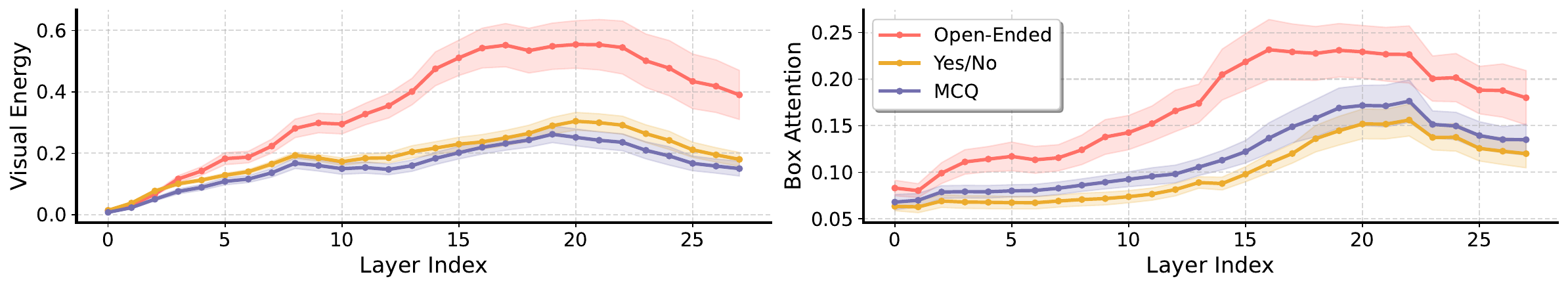}
  
  \vspace{-4mm}
  \caption{\textbf{Visual energy drops significantly on non-open-ended framings.} (Top) There is a significant drop in attention applied to the portion of the image containing the object of interest, and a corresponding increase in attention to sink tokens, for yes/no and MCQ framing. (Bottom) Per-layer values of visual energy and box attention of Qwen2.5-VL-7B on GQA$^\text{F}$.}
  \label{fig:combined_analysis}
  \vspace{-6mm}
\end{figure}

\subsection{Framing Reshapes Visual Attention (\textbf{F$\rightarrow$A})}
With the analysis framework built, we examine how question framing affects the model's visual attention strategy. We characterize this along three dimensions: the overall degree of visual engagement, the spatial allocation of attention relative to task-relevant regions, and the dispersion of attention across the image.

\subsubsection{Visual energy and spatial allocation.}
% If the \textbf{F$\rightarrow$A} pathway exists, we should observe measurable changes in both how much and where the model attends to the image under different framings. 
As shown in~\cref{fig:combined_analysis} (top), constrained framings consistently exhibit lower overall visual energy compared to open-ended generation across both GQA$^\text{F}$ and V$^\text{*F}$, indicating reduced reliance on visual content. Beyond this overall reduction, the spatial distribution of attention shifts dramatically. As reported in~\cref{fig:combined_analysis} (top), attention on sink tokens, positions with low semantic relevance identified in prior work~\cite{kang2025see}, increase for both Yes/No and MCQ. In contrast the attention within target area (Box attention) drops from $19\%$ on Open-ended to around $12\%$ on Yes/No and $13\%$ on MCQ, corresponding to a relative drop of $40\%$ on GQA$^\text{F}$. This distribution is even more different on V$^\text{*F}$ which requires higher grounding skill. The relative drop from open-ended to Yes/no or MCQ is roughly $50\%$. The model does not simply attend less to the image; it actively redirects attention away from task-relevant regions toward unrelated regions. Furthermore, the entropy of the attention distribution increases under constrained framings, indicating that the remaining visual attention becomes more diffuse and less focused on any specific region.

\finding{2}{Constrained framings reduce overall visual energy, redirect attention from target regions to unrelated regions, and produce more dispersed attention, confirming that question framing impacts visual attention \textbf{F$\rightarrow$A} in profound ways.}

\subsubsection{Layer-wise analysis}
 Early layers exhibit similar attention patterns across all framings (\cref{fig:combined_analysis}). The divergence emerges in the middle layers (approximately layers 12--22), which prior work~\cite{jiang2025devils} has identified as cross-modal interaction layers, where visual and textual representations are jointly processed. In these layers, both visual energy and bounding box attention drop significantly for Yes/No and MCQ framings compared to open-ended, and this gap persists through the remaining layers.

\subsubsection{Decomposing the framing effect.}
\begin{figure*}[t]
  \centering

  % --- LEFT SIDE: Variations (Previously Right) ---
  \begin{minipage}{0.49\textwidth}
    \centering
    \includegraphics[page=12, trim={200 95 180 106}, clip, width=1\textwidth]{figures/sources/main.pdf}
    \vspace{-4mm}
    \scriptsize
    \resizebox{0.95\linewidth}{!}{%
    \begin{tabular}{lcccc}
      \toprule
      & \multicolumn{2}{c}{\textbf{V$^\text{*F}$}} & \multicolumn{2}{c}{\textbf{GQA$^\text{F}$}} \\
      \cmidrule(lr){2-3} \cmidrule(lr){4-5}
      \textbf{Variation} & \textbf{VE} & \textbf{Box} & \textbf{VE} & \textbf{Box} \\
      \midrule  
      Question & \textbf{0.146} & \textbf{0.171} & \textbf{0.206} & \textbf{0.109} \\
      Instruction & 0.054 & 0.046 & 0.054 & 0.029 \\
      \bottomrule
    \end{tabular}
    }
    \vspace{3mm}
    \caption{(Top) Illustration of question/instruction variation. (Bottom) Coefficient of quartile variation on VE and Box for varying framing and instruction.}
    \label{fig:variation}
  \end{minipage}
  \hfill
  % --- RIGHT SIDE: Correlation Coefficients (Previously Left) ---
  \begin{minipage}{0.49\textwidth}
    \centering
    \includegraphics[page=1, trim={5 -5 -10 0}, clip, width=1\textwidth]{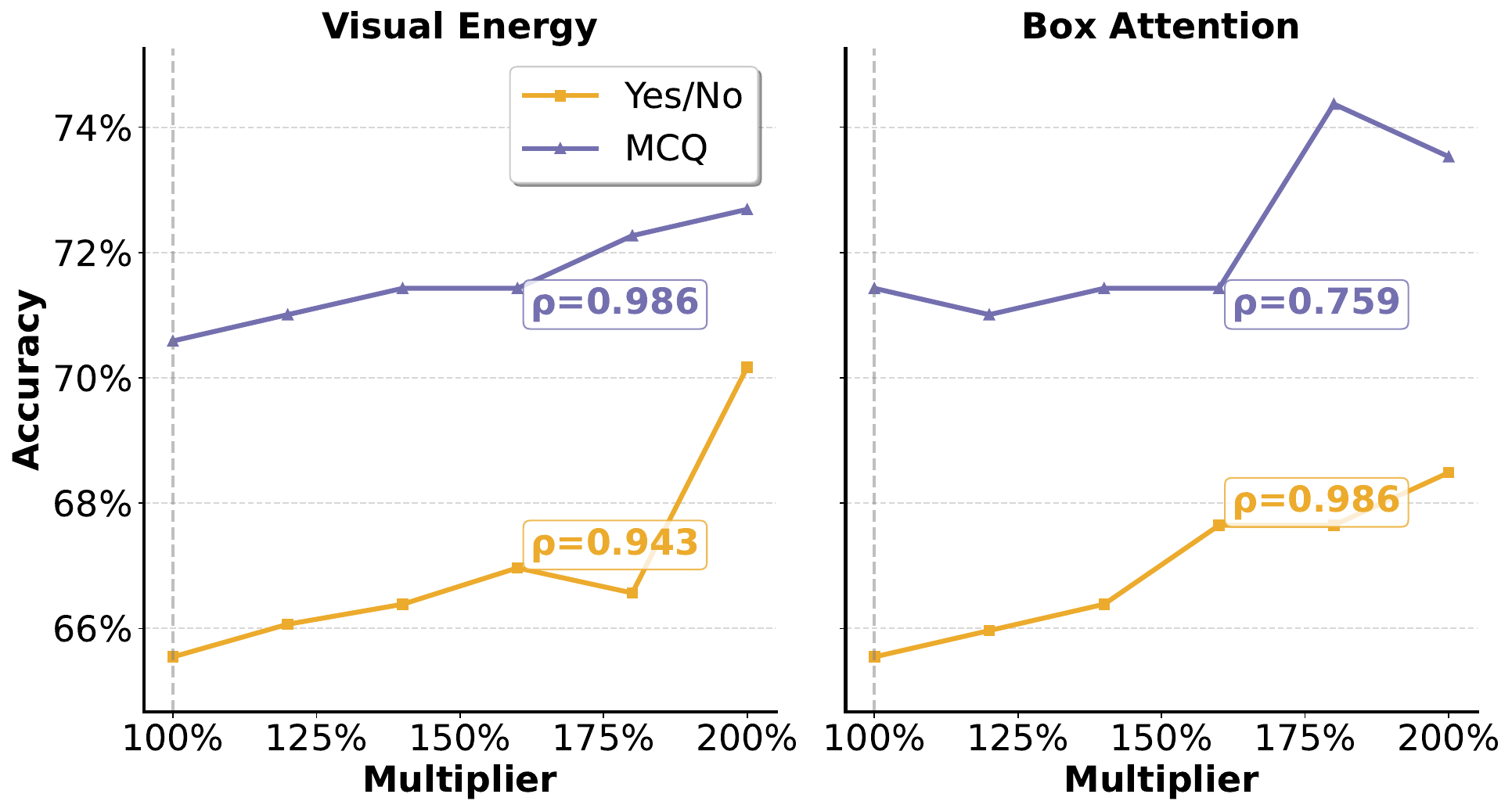}
    \vspace{0mm}
    \scriptsize
    \resizebox{0.95\linewidth}{!}{%
    \begin{tabular}{lcccc}
      \toprule
      & \multicolumn{2}{c}{\textbf{V$^\text{*F}$}} & \multicolumn{2}{c}{\textbf{GQA$^\text{F}$}} \\
      \cmidrule(lr){2-3} \cmidrule(lr){4-5}
      \textbf{Accuracy} & \textbf{Y/N} & \textbf{MCQ} & \textbf{Y/N} & \textbf{MCQ} \\
      \midrule
      Baseline & 0.655 & 0.706 & 0.815 & 0.741 \\
      VE Intervention & \textbf{0.682} & 0.722 & 0.817 & 0.743 \\
      Box Intervention & 0.675 & \textbf{0.735} & \textbf{0.826} & \textbf{0.756} \\
      \bottomrule
    \end{tabular}
    }
    \vspace{0mm}
    \caption{(Top) Performance gains on V$^\text{*F}$ when multiplier increases with Spearman's rank correlation coefficient listed. (Bottom) Attention steering results on V$^\text{*F}$ and GQA$^\text{F}$.}
    \label{fig:correlation}
  \end{minipage}
  \vspace{-6mm}
\end{figure*}

A question prompt consists of two components: the question itself (e.g., ``How many dogs are there?'' vs.\ ``Is there a dog?'') and appended instructions (e.g., ``Answer with Yes or No''). To understand what drives the attention shift, we disentangle these two sources of variation. As illustrated in~\cref{fig:variation} (top), we separately vary the question framing while holding instructions fixed, and vary instructions while holding the question fixed. The coefficient of quartile variation (CQV) for both visual energy and bounding box attention, shown in~\cref{fig:variation} (bottom), reveals that variation from changing the question framing is $\approx3$ times larger than variation from changing instructions alone.

\subsection{Connecting Attention to Prediction (\textbf{A$\rightarrow$Y})}

We now test whether this attention distortion directly drives prediction errors (\textbf{A$\rightarrow$Y}). The correlations observed in previous sections between framing and attention do not imply a direct link to accuracy: the model may simply find constrained framings easier to resolve and naturally allocate less visual energy to target area attention without any cost to performance. In other words, the drop in visual engagement could be an efficient strategy rather than a harmful one. To distinguish these alternatives, we perform \textit{attention steering}, directly intervening on the model's attention maps under constrained framings to restore them toward open-ended levels, and measure whether accuracy recovers.

\subsubsection{Attention steering.}
Ideally, one would copy/intervene the full attention maps from an open-ended forward pass onto a constrained one. In practice, however, question tokens differ across framings, so the attention maps collected have different query dimensions. We therefore intervene on attention properties, magnitude and spatial distribution, using a multiplier-based scheme. We study two complementary interventions that together disentangle the contributions of visual engagement magnitude and spatial allocation:

\paragraph{Visual Energy (VE) steering}
Given a VQA question in different framings, we first compute the ratio of visual energy between open-ended and constrained framings from attention rollout as the multiplier. Then, we perform inference for the constrained framing again: for each head and layer, we scale up the attention weights on all image tokens (for every query token after the image) by this multiplier, while proportionally reducing the attention on non-image tokens to maintain a valid probability distribution. The spatial distribution within image tokens is preserved; only the total visual energy changes, isolating the effect of how much the model attends to the image content.

\paragraph{Box attention steering}
With the task-relevant image regions identified using the ground truth bounding boxes, we compute a separate multiplier as the ratio of bounding-box attention from the open-ended rollout to that of the constrained framing. During inference for the constrained framing, we scale up attention weights of image tokens within the bounding-box by the computed multiplier while keeping the total visual energy constant. This isolates the complementary question: does where the model attends within the image matter?

\subsubsection{Results.}
We report accuracy after steering in~\cref{fig:correlation} (bottom). On V$^\text{*F}$, which demands fine-grained grounding, both interventions yield clear gains: VE steering improves both Y/N and MCQ by $+2.7$ and $+1.6$ pts, while Box steering improves Y/N and MCQ by $+2.0$ and $+2.9$ pts. On GQA$^\text{F}$, a more general reasoning benchmark, VE steering barely helps ($\approx +0.2$ pts for both framings), yet Box steering still delivers consistent improvement ($\approx +1.3$ pts for both). We further visualize performance change via incremental multiplier increases in \cref{fig:correlation} (top) on V$^\text{*F}$. We show that accuracy improves monotonically as attention is steered closer to open-ended levels, with high Spearman correlations. Together, these results indicate that spatial misallocation, where attention falls within the image, matters more universally than total visual energy, especially for grounding-heavy tasks where precise localization is critical.

\vspace{2mm}
\finding{3}{Attention steering confirms that changes induced by framing in the attention directly impact model output \textbf{A$\rightarrow$Y}: restoring attention under constrained framings recovers accuracy. Spatial allocation (where the model looks) yields universal gains, while visual energy magnitude (how much it looks) primarily benefits grounding-heavy tasks, revealing that framing-induced errors stem from qualitatively different attention failures depending on the task.}

\begin{figure*}[t]
  \centering
  \includegraphics[page=11, trim={10 80 10 60}, clip, width=\textwidth]{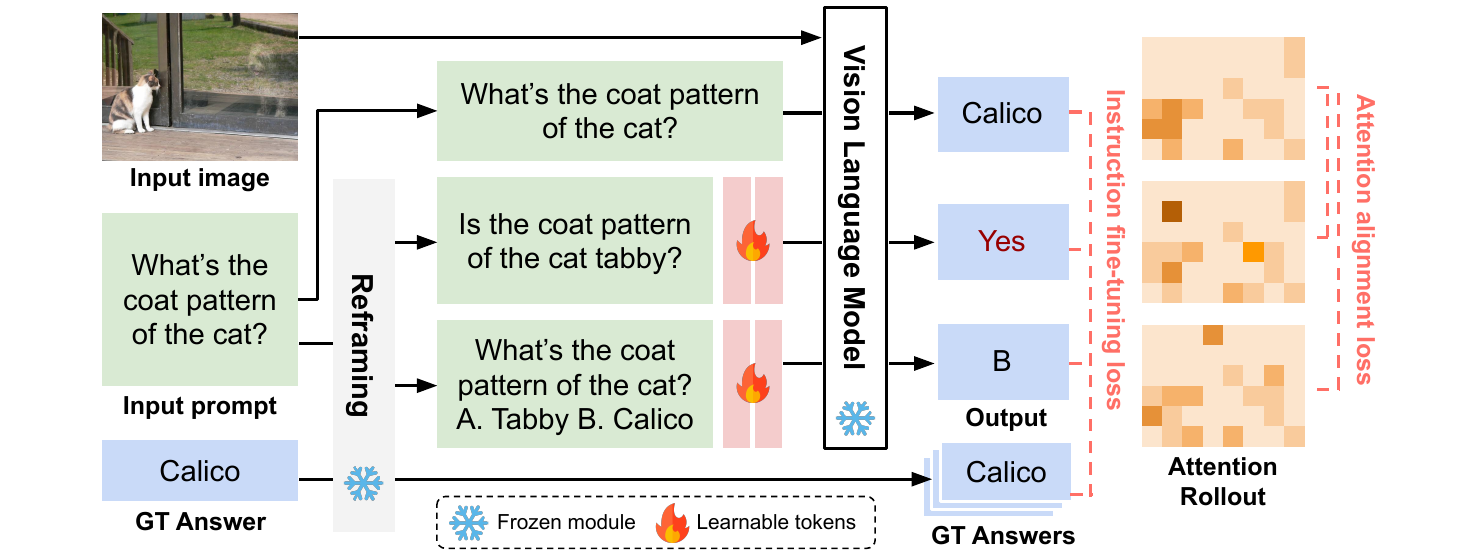}
  \vspace{-2mm}
  \caption{\textbf{Learning to re-align attention}. Given a VQA sample, we reframe the question/answer pairs into alternate framing with a frozen LLM. We append sets of learned tokens to constrained framings at the end of the input sequence. During training, we add an attention alignment objective on the attention rollout to align the attention from open-ended samples with constrained framings.}
  \label{fig:framework}
\end{figure*}   

\subsubsection{Takeaways.}
Taken together, Sections~3--4 establish a three-link pathway from question framing to model failure. First, constrained framings (Y/N, MCQ) can cause prediction errors, with the largest inconsistency happening on grounding-heavy tasks where precise visual localization is essential (\textbf{Finding~1}: \textbf{F$\rightarrow$Y}). Second, we further find that constrained framings reduce overall visual energy, redirect engagement away from task-relevant regions (\textbf{Finding~2}: \textbf{F$\rightarrow$A}). Third, directly steering visual attention back to open-ended levels recovers accuracy, confirming that spatial allocation matters universally while magnitude primarily benefits grounding tasks (\textbf{Finding~3}: \textbf{A$\rightarrow$Y}). Combining these findings clearly shows that framing-induced errors are not superficial, but are induced by the fundamental shift of how visual information is processed by the VLM.
% grounded to qualitative and quantitative failures. 
In the next section, armed with these insights, we propose a simple mitigation strategy that improves robustness across question framings.

% \vspace{-2mm}
\section{Attention Realignment via Prompt Tuning}
\label{sec:methodology}

The above observations that constrained framings suppress visual energy and redirect attention away from task-relevant regions, yet steering attention back recovers accuracy, lead to a natural hypothesis regarding mitigation. Specifically, (1) we hypothesize that the open-ended attention pattern, where the model reasons correctly, can serve as a reliable supervisory signal for realigning constrained framings. Further, (2) since the failure originates at the prompt level rather than from a fundamental model deficiency, lightweight prompt tuning with a small number of learnable tokens should be able to restore proper attention behavior without modifying any model weights.

\begin{figure*}[t]
  \centering
  \includegraphics[page=1, trim={0 0 0 0}, clip, width=\textwidth]{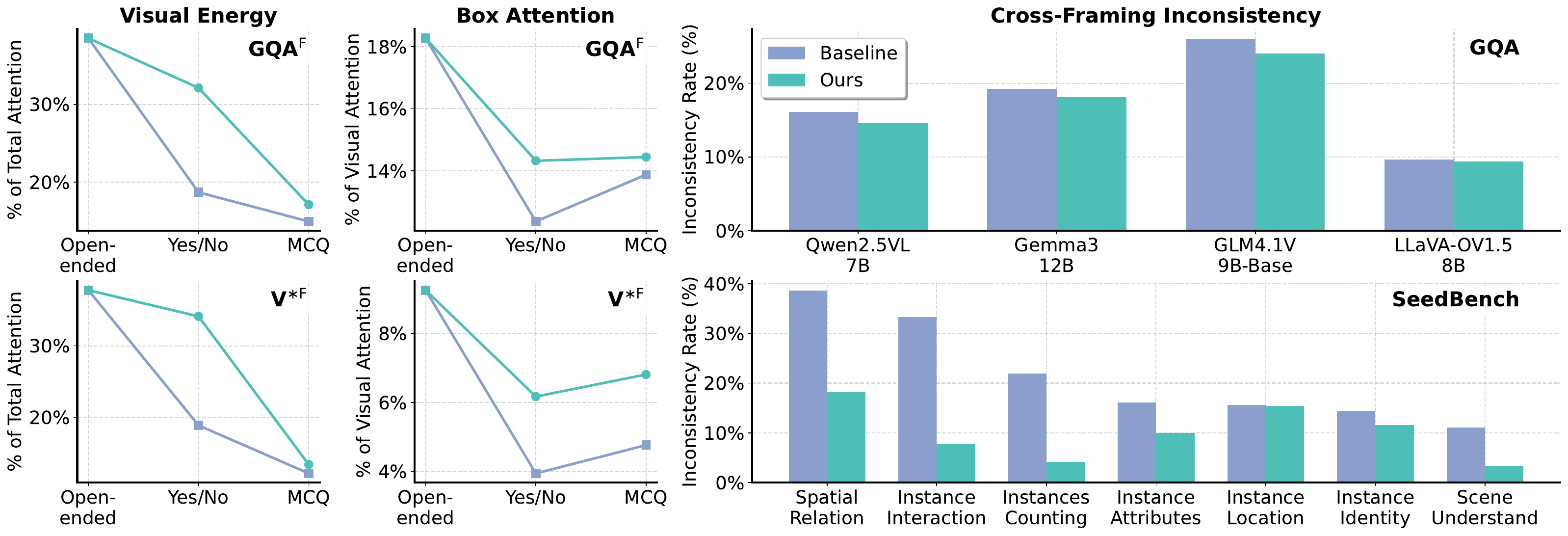}
  \caption{(Left) Performance comparison after appending ours learned soft tokens on input prompts on GQA$^\text{F}$ and V$^\text{*F}$. (Right) Cross-framing inconsistency comparison between baseline (Qwen2.5-VL-7B) and ours.}
  \label{fig:comparison}
\end{figure*}

\subsubsection{Realigning attention via prompt tuning.}
As illustrated in \cref{fig:framework}, our method operates on training triplets constructed from each sample. Given an image and a question with a ground-truth answer, we use Qwen3-32B~\cite{bai2025qwen3} to rewrite the question into all three framings, including open-ended, yes/no, multiple-choice, along with their corresponding answers. For the yes/no and multiple-choice framings, we append $K$ learnable tokens to the input sequence; the open-ended input remains unchanged. All three framings are fed into a frozen VLM in parallel to produce output logits and attention rollouts.

Once the outputs are obtained, we jointly optimize two objectives. The first one is a standard cross-entropy loss for next-token prediction, applied to the triplets, preserving the original question answering capability. The second is an attention alignment loss that encourages the constrained framings to mimic the open-ended attention pattern. Inspired by the attention distillation objective~\cite{fuller2025lookwhere}, our alignment loss operates on the attention rollouts and consists of (1) an $L_2$ loss on visual energy magnitude, penalizing the difference in total attention over image tokens; and (2) a KL divergence on the normalized visual attention distributions (summing to one), encouraging spatial alignment. The alignment loss is computed between the open-ended framing and both constrained framings.

% \vspace{-2mm}
\subsubsection{Implementation and training details.}
We fine-tune the VLM on $10$K randomly sampled VQA pairs from the LLaVA~\cite{liu2023visual} instruction tuning set\footnote{\href{https://huggingface.co/datasets/kaiyuyue/llava-1.5-665k-instructions}{llava-1.5-665k-instructions}}. The model is trained for $1$ epoch using the AdamW optimizer~\cite{loshchilov2017decoupled} with $\beta_1 = 0.9$ and $\beta_2 = 0.999$. We employ a peak learning rate of $2E-4$ with a $5\%$ linear warmup, followed by a cosine decay schedule. We use a batch size of $1$ with gradient accumulation over $16$ steps.
We weight each sample by the model's own confidence in the ground-truth answer under teacher forcing. Specifically, we compute the average probability assigned to each ground-truth token at the corresponding position, yielding a continuous weight. Samples where the model is already confident in the correct answer produce attention patterns that are more likely to be reliable supervisory signals, and thus receive higher weight; samples where the model is uncertain contribute proportionally less. We used $K=8$ learned tokens during fine-tuning. We study the effect of the loss components and the number of learnable tokens $K$ in our ablation experiments (see \cref{sec:exp_results}).

% \vspace{-2mm}
\subsection{Experiments}
\label{sec:exp_results}

% \vspace{-2mm}
\subsubsection{Models and benchmarks.}
We evaluate our method on five VLMs spanning diverse architectures: Qwen2.5-VL-7B~\cite{bai2025qwen25vltechnicalreport}, Qwen3-VL-8B~\cite{bai2025qwen3}, LLaVA-OneVision-1.5-8B~\cite{an2025llava}, Gemma3-12B~\cite{team2025gemma}, and GLM4.1V-9B-Base~\cite{zeng2025glm}. For each model, the learnable tokens are trained on 10K randomly sampled examples (with early stopping) from the LLaVA SFT training set, as described above. We evaluate downstream performance on 7 benchmarks grouped into three categories: general and reasoning (RealWorldQA$^\dagger$~\cite{xai2024grok15v}, MME~\cite{fu2023mme}, MMMU-Pro$^\ddagger$~\cite{yue2025mmmu}), alignment (HallusionBench$^{\ddagger}$~\cite{guan2024hallusionbench}, POPE~\cite{li2023evaluating}), and fine-grained grounding (HRBench8k~\cite{wang2025divide}, V$^*$~\cite{wu2024v}). We note that MME and POPE are yes/no benchmarks, while the rest are multiple-choice QAs. Since we focus on constrained framings, we note that $^\dagger$ denotes the multiple-choice subset, and $^\ddagger$ denotes the vision split of MMMU-Pro and HallusionBench. Also, we do not incorporate the tools used for VLMs when testing on V$^*$. In addition, we revisit the GQA, Seedbench, GQA$^\text{F}$, and V$^{\text{*F}}$ evaluation setup from \cref{sec:analysis} to measure whether our method recovers visual attention and reduces cross-framing inconsistencies. 

\begin{table*}[t]
\centering
\setlength{\tabcolsep}{3pt}
\renewcommand{\arraystretch}{1}
\resizebox{0.95\columnwidth}{!}{
\begin{tabular}{l|ccc|cc|cc}
\toprule
\multirow{3}{*}{\textbf{Methods}} & \multicolumn{3}{c|}{\textbf{General \& Reasoning}} & \multicolumn{2}{c|}{\textbf{Alignment}} & \multicolumn{2}{c}{\textbf{Fine-grained grounding}} \\
\cmidrule(lr){2-4} \cmidrule(lr){5-6} \cmidrule(lr){7-8}
& RealWorldQA$^\dagger$ & MME & MMMU-Pro$^\ddagger$ & Hallusion$^\ddagger$ & POPE & HRBench8k & V$^\text{*}$ \\
\midrule
Gemma3-12B & \textbf{61.87} & 2133.5 & 27.92 & 64.56 & 84.63 & 48.25 & 48.50 \\
\rowcolor{myblue}
+ Ours & \textbf{61.87} & \textbf{2180.0} & \textbf{28.09} & \textbf{66.14} & \textbf{84.98} & \textbf{49.50} & \textbf{49.79} \\
\midrule
GLM4.1V-9B-Base & \textbf{69.41} & 2208.1 & \textbf{32.22} & 63.72 & \textbf{88.52} & 51.62 & 58.80 \\
\rowcolor{myblue}
+ Ours & \textbf{69.41} & \textbf{2237.8} & 31.41 & \textbf{67.51} & 87.24 & \textbf{52.25} & \textbf{61.80} \\
\midrule
LLaVA-OV1.5-8B & \textbf{70.09} & 2276.2 & 30.52 & 66.04 & 88.82 & 58.00 & 62.23 \\
\rowcolor{myblue}
+ Ours & 69.63 & \textbf{2290.3} & \textbf{30.92} & \textbf{66.14} & \textbf{89.12} & \textbf{58.50} & \textbf{62.66} \\
\midrule
Qwen2.5VL-7B & 63.70 & 2199.5 & 31.45 & \textbf{71.50} & 87.68 & \textbf{54.13} & 66.95 \\
\rowcolor{myblue}
+ Ours & \textbf{66.44} & \textbf{2269.7} & \textbf{31.50} & 68.77 & \textbf{88.65} & \textbf{54.13} & \textbf{69.53} \\
\midrule
Qwen3VL-8B & 70.78 & 2390.8 & 32.72 & 71.08 & 89.26 & \textbf{61.38} & 72.96 \\
\rowcolor{myblue}
+ Ours & \textbf{71.92} & \textbf{2430.5} & \textbf{32.89} & \textbf{72.34} & \textbf{89.41} & 61.25 & \textbf{75.11} \\
\bottomrule
\end{tabular}}
\vspace{1mm}
\caption{Performance impact of our method. $^\dagger$ denotes the multiple-choice subset, and $^\ddagger$ denotes the vision split.}
\vspace{-6mm}
\label{tab:main}
\end{table*}

% \vspace{-2mm}
\subsubsection{Improving visual attention and prediction consistency across framings.}
Before investigating downstream accuracy, we first verify the effectiveness of our proposed method on restoring the attention for constrained framings. As the results shown in~\cref{fig:comparison}~(Left), after applying our learnable tokens on Yes/No and MCQ, visual energy under constrained framings recovers substantially and bounding box attention increases, indicating that the model redirects its focus back to task-relevant image regions. We also revisit the inconsistency evaluation in~\cref{fig:comparison}~(right). One can see that the cross-framing inconsistency rate drops generally across all VLMs on GQA. On the Seedbench task breakdown using Qwen2.5-VL-7B, we observe significant inconsistency reductions by $20\%$ on instance interaction and $15\%$ on counting categories that demand fine-grained visual grounding. These results confirm that the learnable tokens successfully realign the attention distribution of constrained framings toward the open-ended pattern, motivating the benchmark evaluation that follows.

% \vspace{-2mm}
\subsubsection{Improving the performance across benchmarks.}
We evaluate whether re-aligned attention translates into downstream accuracy gains, as hypothesized in~\cref{sec:hypothesis}. We append learned tokens to the input sequence according to the framing type of each benchmark (i.e., yes/no or MCQ). In~\cref{tab:main}, we report results across $7$ benchmarks and $5$ VLMs spanning 7B to 12B. Overall, our methods yields consistent improvements across most model and benchmark combinations. The gains are most pronounced on fine-grained grounding tasks: on V$^*$, Qwen2.5-VL-7B improves by $2.5$pp, while HRBench8k sees steady improvements for Gemma3 and GLM4.1V. The results resonate the findings that tasks demanding spatial localization benefit most from restored visual attention. General reasoning benchmarks show modest but positive gains overall. The consistency of the improvements models confirms the effectiveness of our methods.

% \vspace{-4mm}
\subsubsection{Model ablation.}
To verify that the performance gains stem from attention alignment rather than additional instruction tuning with cross-entropy loss, we first ablate the loss components on Qwen2.5-VL-7B. The results are shown in~\cref{tab:ablation_combined}. Removing the attention alignment loss yields marginal gains over the baseline, indicating that cross-entropy loss alone is insufficient and barely recovers the performance of the baseline after adding additional learnable tokens; explicit attention realignment is the primary driver of improvement. Furthermore, removing the cross-entropy loss causes significant performance drops, as the additional learnable tokens are initialized randomly and the CE loss is essential to keep the model's question-answering capability functioning normally. The full model achieves the best results across most metrics.

\begin{table*}[t]
\centering
\scriptsize
\renewcommand{\arraystretch}{0.8}
% --- Left Table: Loss Ablation ---
\begin{minipage}{0.5\textwidth}
\centering
% \caption{\textbf{Ablation on loss functions.}}
\resizebox{1\columnwidth}{!}{
\begin{tabular}{l|ccccc}
\toprule
\textbf{Ablation} & \textbf{RWQA$^\dagger$} & \textbf{MMMU-Pro$^\ddagger$} & \textbf{POPE} & \textbf{V$^\text{*}$} & \textbf{Avg} \\
\midrule
Baseline & 63.70 & 31.45 & 87.68 & 66.95 & 62.45 \\
\rowcolor{myblue}
Ours & \textbf{66.44} & 31.50 & \textbf{88.65} & \textbf{69.53} & \textbf{64.03} \\
w/o Attn Loss & 64.84 & \textbf{32.89} & 87.05 & 65.67 & 62.61 \\
w/o CE Loss & 64.61 & 31.27 & 83.40 & 63.95 & 60.81 \\
\bottomrule
\end{tabular}}
\end{minipage}
\hfill
% --- Right Table: Token Ablation ---
\begin{minipage}{0.48\textwidth}
\centering
\resizebox{1\columnwidth}{!}{
\begin{tabular}{l|ccccc}
\toprule
\textbf{Tokens} & \textbf{RWQA$^\dagger$} & \textbf{MMMU-Pro$^\ddagger$} & \textbf{POPE} & \textbf{V$^\text{*}$} & \textbf{Avg} \\
\midrule
4 & 63.70 & 31.33 & 88.43 & 66.09 & 62.39 \\
\rowcolor{myblue}
8 & \textbf{66.44} & \textbf{31.50} & 88.65 & \textbf{69.53} & \textbf{64.03} \\
12 & 63.93 & \textbf{31.50} & 88.36 & 67.81 & 62.90 \\
16 & 63.93 & 30.92 & \textbf{88.66} & 67.38 & 62.72 \\
\bottomrule
\end{tabular}}
\end{minipage}
\vspace{1mm}
\caption{Ablation on loss functions and learnable token count. RWQA denotes RealWorldQA. $^\dagger$ denotes the multiple-choice subset, and $^\ddagger$ denotes the vision split.}
\label{tab:ablation_combined}
\vspace{-8mm}
\end{table*}
\section{Conclusion}

In this paper, we establish that visual attention within VLMs is significantly impacted by question framing. Through our mechanistic analysis of the processing pathways in a VLM, we establish a latent relationship between framing and downstream task performance. This work reframes visual blindness as a dynamic behavior controllable by a user rather than a static architectural limitation. With this results, we implement a prompt-tuning method that realigns attention under constrained framings, yielding consistent improvement across models and benchmarks without modifying models' weights.

\section*{Acknowledgements}
{\small This work was funded, in part, by the Vector Institute for AI, Canada CIFAR AI Chairs, NSERC Canada Research Chair (CRC), AML-TN UBC, and NSERC Discovery and Discovery Accelerator Supplement Grants. Resources used in preparing this research were provided, in part, by the Province of Ontario, the Government of Canada through CIFAR, the Digital Research Alliance of Canada\footnote{\url{alliancecan.ca}}, companies\footnote{\url{https://vectorinstitute.ai/\#partners}} sponsoring the Vector Institute, and Advanced Research Computing at the University of British Columbia. Additional hardware support was provided by John R. Evans Leaders Fund CFI grant and Compute Canada under the Resource Allocation Competition award. Ritwik Gupta and Declan Kutscher were supported in part by funding from the Department of Defense, The House Fund, and BAIR’s industrial alliance programs. Additional compute was provided by the Department of Defense's High Performance Computing Modernization Program. We are immensely grateful to Bicheng Xu from UBC and Stephanie Fu from UCB for sharing their valuable suggestions in paper writing and experiments.}

\clearpage  % TODO FINAL: This 

\bibliographystyle{splncs04}
\bibliography{main}

% \startcontents[app]
\appendix
% \printcontents[app]{}{1}{}

\clearpage

\setcounter{section}{0}
\setcounter{table}{0}
\setcounter{figure}{0}
\setcounter{page}{1}
\def\thesection{\Alph{section}}
\renewcommand{\thetable}{A\arabic{table}}
\renewcommand{\thefigure}{A\arabic{figure}}

\begin{abstract}
    This supplementary provides additional implementation details, quantitative results, and qualitative analyses supporting the main paper. Specifically, \cref{sec:impl} covers implementation details, including the cross-framing inconsistency pipeline and GPT prompts for question reframing (\cref{sec:impl_inconsistency}), curation details and human evaluation of GQA$^\text{F}$ and V$^{*\text{F}}$ (\cref{sec:impl_curation}), training and resource usage (\cref{sec:impl_training,sec:impl_resource}), and evaluation protocols for all benchmarks (\cref{sec:impl_eval}). \cref{sec:quant} presents additional quantitative results, including ablation studies on learnable token placement and confidence-based loss weighting (\cref{sec:ablation}), and extended visual attention analysis across additional VLM families, Gemma3, GLM-4.1V, LLaVA-OneVision-1.5, and Qwen3VL (\cref{sec:vis_analysis}). \cref{sec:qual} provides additional qualitative results, and \cref{sec:limitation} discusses limitations and future directions.
\end{abstract}

\section{Implementation Details}
\label{sec:impl}

\subsection{Cross-Framing Inconsistency}
\label{sec:impl_inconsistency}
As discussed in~\cref{sec:f2a2y}, we leverage cross-framing inconsistency to verify whether question framing affects the model's final prediction (\textbf{F$\rightarrow$Y} in~\cref{fig:causal_graph}), and the evaluation pipeline is illustrated in~\cref{fig:sec3_inconsistency} (left). Here, we discuss more implementation details. For data curation, SeedBench~\cite{li2023seed} is a MCQ benchmark, so we remove the possible answer choices to obtain the open-ended question and use the content of the correct option as the ground truth answer. GQA~\cite{hudson2019gqa} is largely open-ended, but a small portion of it consists of Yes/No questions. We filter out the Yes/No questions in GQA to obtain a fully open-ended generation benchmark. We then perform inference on both datasets across different VLMs and evaluate the performance. For samples which were answered correctly, we use GPT-5.1 (\texttt{gpt-5.1-2025-11-13}) to reframe the questions into Yes/No or MCQ (while for SeedBench we use the original options), and perform inference on the questions in different framings to get the final inconsistency rate. The detailed system prompt for GPT-5.1 is provided below. We study the quality of question reframing using GPT in the following data curation section.

\begin{promptbox}[title=Prompt for Reframing Open-ended Question to MCQ/YN questions]
\small
Given the following question and its correct answer, create TWO new framings:

\medskip
\textbf{Original Question:} \texttt{\{original\_question\}}\\
\textbf{Correct Answer:} \texttt{\{correct\_answer\}}\\
\textit{[If options exist:]} \textbf{Available Options:} \texttt{\{options\}}

\medskip
\textbf{1. Yes/No:}
\begin{itemize}[noitemsep, topsep=2pt]
  \item \textit{[If affirm:]} Convert to a binary yes/no question where the answer is `yes' (affirming that the answer is \texttt{\{correct\_answer\}}).
  \item \textit{[If negate:]} Convert to a binary yes/no question where the answer is `no'. Replace the correct answer (\texttt{\{correct\_answer\}}) with a distinctly incorrect or unlikely alternative that is clearly contradicted by the original answer. Avoid any ambiguous or `near-miss' alternatives. Keep the same target object that the original question refers to if possible --- do not change the subject or object being asked about.
  \item Turn the original question into a yes/no format that tests the same knowledge. For example, if the original question is ``What is the person near the garbage bin wearing?'' with answer ``a coat'', the yes/no question could be ``Is the person near the garbage bin wearing a coat?''
  \item Must be a binary question (Is/Are/Does/Do/Can/Could/etc.)
\end{itemize}

\medskip
\textbf{2. MCQ (Multiple Choice Question):}
\begin{itemize}[noitemsep, topsep=2pt]
  \item \textit{[If options exist:]} Create an MCQ version with the same options: \texttt{\{options\}}. The correct answer should be \texttt{\{correct\_answer\}}.
  \item \textit{[If no options:]} Create an MCQ version with 4 options. The correct answer should be \texttt{\{correct\_answer\}}, and provide 3 easy negative distractors.\\
  \textit{[If scene graph available:]} Avoid using objects from the scene graph as negative options as they may create ambiguity. Scene graph: \texttt{\{scene\_graph\}}.
  \item Should be same open-ended question (What/Which/Where/Who/How/etc.)
  \item Provide exactly 4 options as a list
\end{itemize}

\medskip
\textbf{IMPORTANT RULES:}
\begin{itemize}[noitemsep, topsep=2pt]
  \item Yes/No question must test the SAME knowledge as the original
  \item MCQ must test the SAME knowledge as the original
  \item Both should be answerable from the same visual information
\end{itemize}

\medskip
Output ONLY valid JSON in this exact format:
\begin{verbatim}
{
  "yes_no": {
    "question": "Is/Are/Does/Do... question",
    "answer": "{yes|no}"
  },
  "mcq": {
    "question": "What/Which/Where... question",
    "options": ["option1", "option2", "option3", "option4"],
    "answer_text": "{correct_answer}"
  }
}
\end{verbatim}
\end{promptbox}

\begin{promptbox}[title=Prompt for Reframing YN Question to Open-ended/MCQ questions]
\small
Given the following yes/no question and answer, create TWO new framings:

\medskip
\textbf{Original Question:} \texttt{\{original\_question\}}\\
\textbf{Correct Answer:} \texttt{\{original\_answer\}}\\
\textbf{Correct Full Answer} (for reference)\textbf{:} \texttt{\{original\_full\_answer\}}

\medskip
\textbf{1. MCQ (Multiple Choice Question):}
\begin{itemize}[noitemsep, topsep=2pt]
  \item Convert to a WH-question (What/Which/Where/Who/How/etc.) that asks for identification
  \item Provide 4 plausible answer options as a list (the correct answer should be one of them and the answer should not be yes/no)
  \item Example: ``What color is the helmet?'' with options [``light blue'', ``red'', ``yellow'', ``green'']
  \item \textit{[If scene graph available:]} Avoid using objects from the scene graph as negative options as they may create ambiguity. Scene graph: \texttt{\{scene\_graph\}}.
\end{itemize}

\medskip
\textbf{2. Open-ended (short answer):}
\begin{itemize}[noitemsep, topsep=2pt]
  \item Convert to a WH-question (What/Which/Where/Who/How/etc.) expecting a brief specific answer
  \item Should ask the same thing as MCQ but without providing options
  \item Example: ``What color is the helmet in the middle?'' $\rightarrow$ Answer: ``light blue''
\end{itemize}

\medskip
\textbf{IMPORTANT RULES:}
\begin{itemize}[noitemsep, topsep=2pt]
  \item MCQ and Open-ended should use WH-questions (What/Which/Where/Who/How/When/Why) and the answer should not be yes/no
  \item Both formats should test the SAME knowledge as the original question
\end{itemize}

\medskip
Output ONLY valid JSON in this exact format:
\begin{verbatim}
{
  "mcq": {
    "question": "What/Which/Where... question",
    "options": ["option1", "option2", "option3", "option4"],
    "answer_text": "<correct answer>"
  },
  "open_ended": {
    "question": "What/Which/Where... question",
    "answer": "<correct answer>"
  }
}
\end{verbatim}
\end{promptbox}

\subsection{Curation Details of GQA$^\text{F}$, and V$^\text{*F}$}
\label{sec:impl_curation}
As mentioned in~\cref{sec:analysis}, to isolate the impact of task framing from variations in question content, we curate two datasets where each sample has three distinct framing variants: open-ended, Yes/No, and MCQ. We ensure that the underlying visual reasoning required remains constant while only the output format changes. Again, we utilize GPT-5.1 to rephrase the original samples into these target formats. Now, we discuss how we curate datasets including GQA$^\text{F}$ and V$^\text{*F}$ in detail and use human evaluation to further verify their quality.

GQA contains both open-ended and yes/no questions. We leverage the ground-truth scene graph when prompting GPT to ensure that generated MCQ distractors are reasonable without requiring visual access. We provide the prompt that we used for reframing.

For V$^\text{*}$, a multiple-choice benchmark, we remove the answer choices to obtain a purely open-ended question, and then follow the same protocol as used in GQA to reframe it into a Yes/No question.

After prompting and reframing, we filter out samples not in the correct JSON format. We curate a final dataset of 10k unique semantic queries for GQA and the full 300 samples for V$^*$. With three framing variants per query, we obtain 30k and 900 samples for GQA and V$^*$, respectively.

\subsubsection{Human evaluation}
\begin{figure*}[t]
  \centering
  \begin{minipage}{0.48\textwidth}
    \centering
    \includegraphics[page=13, trim={80 100 80 100}, clip, width=\textwidth]{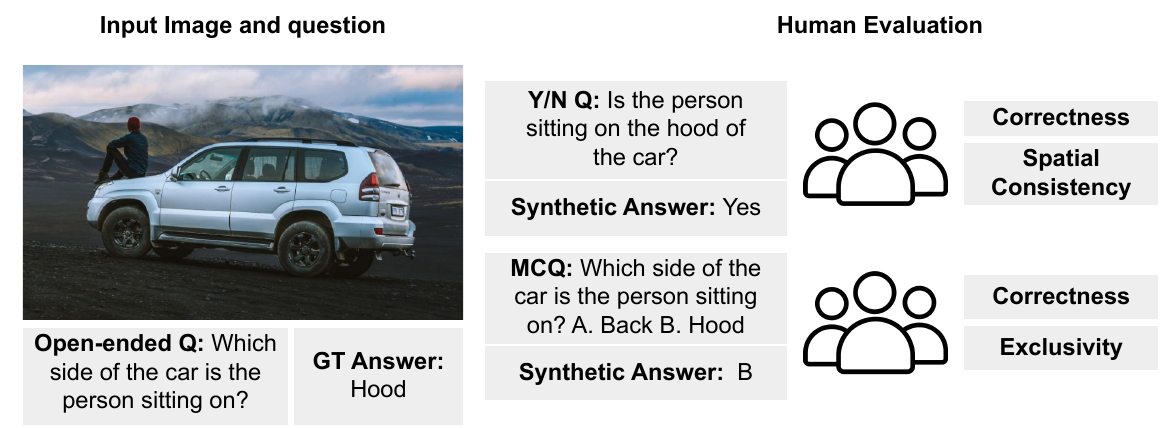}
  \end{minipage}
  \hfill
  \begin{minipage}{0.48\textwidth}
    \centering
    \includegraphics[page=14, trim={80 100 80 100}, clip, width=\textwidth]{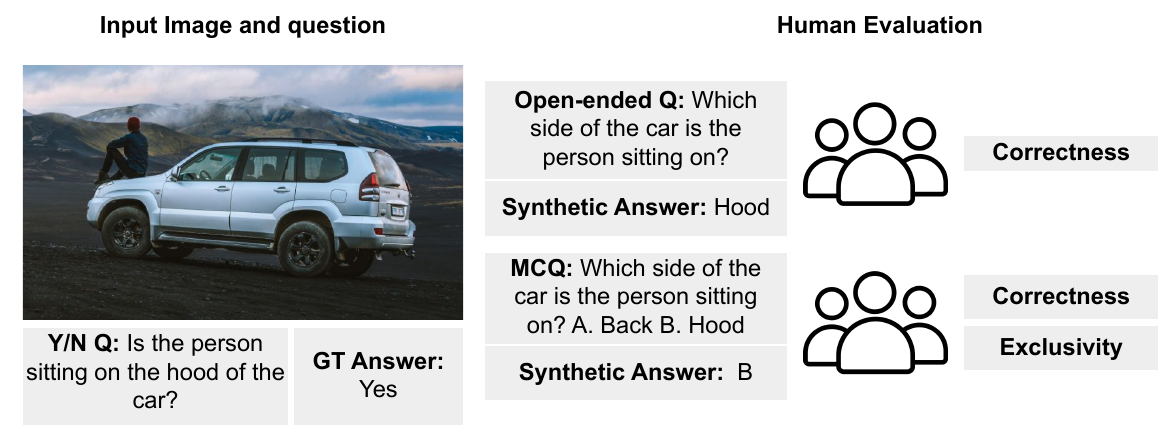}
  \end{minipage}
  
  \caption{Human evaluation pipeline for open-ended and Yes/No question reframing. Note that for MCQ question reframing, we can use a rule-based approach (simply remove the options) to convert it into open-ended and then perform reframing to Yes/No.}
  \label{fig:human_demo}
\end{figure*}
To verify the quality of the reframed questions and answers, we conduct a human evaluation, with results shown in \cref{tab:human_eval}. Recall that the dataset reframing process covers three settings: (1) open-ended $\rightarrow$ MCQ and Yes/No, (2) Yes/No $\rightarrow$ open-ended and MCQ, and (3) MCQ $\rightarrow$ open-ended and Yes/No. Since an MCQ without its answer options is naturally open-ended, we handle this direction via a simple rule-based method that strips the options from the question. The resulting open-ended question can then be further converted to Yes/No using the same pipeline. This reduces the number of conversion types that require evaluation to four: open-ended $\rightarrow$ Yes/No, open-ended $\rightarrow$ MCQ, Yes/No $\rightarrow$ open-ended, and Yes/No $\rightarrow$ MCQ.

We now describe the human evaluation criteria for each of the four conversions. The full human evaluation interface is shown in~\cref{fig:human_demo}.

\noindent\textbf{Open-ended $\rightarrow$ Yes/No.} Annotators assess both answer correctness and \textit{spatial consistency}---whether the target object or region referred to in the question remains the same after reframing. For instance, if an open-ended question asks ``What color is the cake?'' but the reframed Yes/No question asks ``Is the candy purple?'', the target object has shifted, which violates our requirement that the core concept of the question be preserved across formats.

\noindent\textbf{Open-ended $\rightarrow$ MCQ.} Annotators assess answer correctness and additionally evaluate \textit{option exclusivity}---whether the designated correct option is unambiguously the best choice among all provided options. This is necessary because an LLM may inadvertently generate multiple plausible correct answers, introducing ambiguity.

\noindent\textbf{Yes/No $\rightarrow$ Open-ended and Yes/No $\rightarrow$ MCQ.} For both conversions, annotators assess answer correctness. For MCQ, we also evaluate option exclusivity as aforementioned.

The paired human evaluation results are shown in~\cref{tab:human_eval}. We collect 100 samples with corresponding human feedback for each pair (500 samples in total). We use the CloudResearch platform for human studies, with a total of 197 human evaluators involved, coming from 6 countries, including USA, UK, Ireland, Australia, New Zealand, and Canada, whose native language is English. Across all conversion directions, correctness scores remain consistently high, demonstrating that the reframing process is robust. Beyond correctness, we evaluate two format-specific properties: \textit{spatial consistency}, which measures whether target object or task-relevant region remains unchanged after reframing even the generated answer is correct, and \textit{exclusivity}, which measures whether the correct answer in a reframed MCQ question remains unambiguously correct (i.e., the correct option is the best valid option be considered). Both properties are well-maintained, with spatial consistency scores of $96.9\%$ and $95.4\%$ and exclusivity scores of $93.7\%$ and $95.7\%$.

\begin{table}[t]
\centering
\label{tab:human_eval}
\resizebox{0.8\linewidth}{!}{%
\begin{tabular}{lcccccc}
\toprule
\multirow{2}{*}{\textbf{From $\downarrow$ / To $\rightarrow$}} 
  & \multicolumn{1}{c}{\textbf{Open-ended}} 
  & \multicolumn{2}{c}{\textbf{Yes/No}} 
  & \multicolumn{2}{c}{\textbf{MCQ}} \\
\cmidrule(lr){2-2} \cmidrule(lr){3-4} \cmidrule(lr){5-6}
  & Correctness & Correctness & Spatial Cons. & Correctness & Exclusivity \\
\midrule
\textbf{Open-ended} & ---                 & 96.5 & 96.9 & 94.8 & 95.7 \\
\textbf{Yes/No}     & 92.2                & ---  & ---  & 94.8 & 93.7 \\
\textbf{MCQ}        & \textit{rule-based} & 92.6 & 95.4 & ---  & ---  \\
\bottomrule
\end{tabular}%
}
\vspace{1mm}
\caption{Human evaluation results for question reframing across different conversion types. 
MCQ $\rightarrow$ Open-ended is handled via rule-based conversion and requires no human evaluation.}
\end{table}

\subsection{Training}
\label{sec:impl_training}
Following the discussion in~\cref{sec:methodology}, we fine-tune the VLMs on $10$K randomly sampled VQA pairs from the LLaVA~\cite{liu2023visual} instruction tuning set. 
The model is trained for $1$ epoch using the AdamW optimizer~\cite{loshchilov2017decoupled} 
with $\beta_1 = 0.9$ and $\beta_2 = 0.999$. We employ a peak learning rate of $2 \times 10^{-4}$ with a $5\%$ linear warmup, followed by a cosine decay schedule. 
We use a batch size of $1$ with gradient accumulation over $16$ steps. All models, except for GLM, are trained on a single NVIDIA L40. GLM is trained on a single 80GB H100-SXM.

We weight each sample by the model's own confidence in the ground-truth answer under teacher forcing~\cite{vaswani2017attention}. Specifically, we compute the average probability assigned to each ground-truth token at the corresponding position, yielding a continuous weight. Samples where the model is already confident in the correct answer produce attention patterns that are more likely to be reliable supervisory signals, and thus receive higher weight; samples 
where the model is uncertain contribute proportionally less. During training, we jointly apply cross-entropy loss and attention alignment loss. Since the ranges of attention alignment from both mass and distribution are relatively small, we scale the attention loss by a factor of $5$ to match the range of the cross-entropy loss. During training, the maximum image size is limited to $728$ (longest side) for efficiency while the aspect ratio remains unchanged. For the reframing module in the training pipeline, we use Qwen3-32B locally. Gradient checkpointing is disabled for all models since the gradient backpropagation on the attention map may be 
incorrect when the gradient checkpointing is on in PyTorch. Training is done with bf16 and FlashAttention~\cite{shah2024flashattention}. We use $K=8$ learned tokens during fine-tuning.

\subsection{Resource Usage}
\label{sec:impl_resource}
We report the training cost and parameter analysis of our methods in~\cref{tab:resource_usage}. We are only training the learnable tokens for each framing. With $K=8$, the trainable parameters are relatively low at around $60$K. We train all models using one epoch with early-stopping applied, and the training time cost from smallest model at around three hours to largest model to six hours on one single GPU (L40 or H100). VRAM usage is high as the gradient graph are stored backpropogation back to input embedding space.

\begin{table}[h!]
\centering
\scriptsize
\begin{tabular}{@{}lcccc@{}}
\toprule
% --- Training Phase Section ---
& \makecell[c]{\textbf{Trainable} \\ \textbf{Parameters}} & \makecell[c]{\textbf{Time} \\ (min)} & \makecell[c]{\textbf{VRAM} \\ (peak)} & \makecell[c]{\textbf{VIRT} \\ (peak)} \\
\midrule
Qwen2.5VL-7B & 57 K & 187 & 33.8 GB & 53.1 GB \\
Qwen3VL-8B & 66 K & 190 & 42.7 GB & 65.9 GB \\
Gemma3-12B & 61 K & 380 & 39.6 GB & 59.7 GB \\
GLM-4.1V-8B-Base & 66 K & 332 & 76.9 GB & 116.1 GB \\
LLaVA-OneVision-1.5-8B & 66 K & 259 & 36.5 GB & 54.6 GB \\
\bottomrule
\end{tabular}
\vspace{1mm}
\caption{\textbf{Resource usage statistics for prompt tuning.} Tested on 1 NVIDIA L40 with batch size 1 and gradient accumulation 16. VIRT stands for virtual memory size. GLM-4.1V-8B-Base is tested on 1 NVIDIA H100.}
\label{tab:resource_usage}
\end{table}

\subsection{Evaluations}
\label{sec:impl_eval}

\subsubsection{GQA}~\cite{hudson2019gqa} is a general visual question answering benchmark containing open-ended and yes/no questions. When evaluating the score, we follow the official script and perform simple string matching for both yes/no and open-ended questions.

\subsubsection{SeedBench}~\cite{li2023seed} is a comprehensive benchmark containing video question answering, single-image, and multi-image question answering tasks. In this paper, we focus on the single-image task and subsample the single-image portion accordingly. For evaluation, SeedBench is MCQ-type, and we follow the official setting using an answer ranking strategy to obtain the final predicted option. Specifically, we compute the likelihood of each option instead of appending an instruction at the end of the question asking the model to return the letter of the correct option. This evaluation method disentangles the model's instruction-following capability from its ability to answer the given questions.

\subsubsection{RealWorldQA}~\cite{xai2024grok15v} is released by xAI alongside Grok-1.5 Vision and evaluates basic real-world spatial understanding. Each question is appended with options and an instruction for controlling the output format. RealWorldQA is a mixture of MCQ and open-ended questions. As we are testing the improvement of our soft tokens on yes/no and MCQ questions while leaving open-ended question inference untouched, we use only the MCQ portion of the benchmark. We directly use the official questions as input and apply string matching between the ground-truth letter and the predicted letter.

\begin{table*}[t]
\centering
\setlength{\tabcolsep}{3pt}
\renewcommand{\arraystretch}{1}
\resizebox{0.95\columnwidth}{!}{
\begin{tabular}{l|ccc|cc|cc}
\toprule
\multirow{3}{*}{\textbf{Position}} & \multicolumn{3}{c|}{\textbf{General \& Reasoning}} & \multicolumn{2}{c|}{\textbf{Alignment}} & \multicolumn{2}{c}{\textbf{Fine-grained grounding}} \\
\cmidrule(lr){2-4} \cmidrule(lr){5-6} \cmidrule(lr){7-8}
& RealWorldQA$^\dagger$ & MME & MMMU-Pro$^\ddagger$ & Hallusion$^\ddagger$ & POPE & HRBench8k & V$^\text{*}$ \\
\midrule
Prefix & 61.64 & \textbf{2272.2} & 28.15 & 65.51 & 88.28 & 51.62 & 66.52 \\
\rowcolor{myblue}
Infix  & \textbf{66.44} & 2269.7 & \textbf{31.50} & \textbf{68.77} & \textbf{88.65} & 54.13 & \textbf{69.53} \\
Postfix & 63.47 & 2262.4 & \textbf{31.50} & 65.83 & 88.57 & \textbf{54.25} & 67.38 \\
\bottomrule
\end{tabular}}
\vspace{1mm}
\caption{Ablation study on learnable token positioning. $^\dagger$ denotes the multiple-choice subset, and $^\ddagger$ denotes the vision split.}
\vspace{-6mm}
\label{tab:ablation_position}
\end{table*}

\begin{table*}[t]
\centering
\setlength{\tabcolsep}{3pt}
\renewcommand{\arraystretch}{1}
\resizebox{0.95\columnwidth}{!}{
\begin{tabular}{l|ccc|cc|cc}
\toprule
\multirow{3}{*}{\textbf{Strategy}} & \multicolumn{3}{c|}{\textbf{General \& Reasoning}} & \multicolumn{2}{c|}{\textbf{Alignment}} & \multicolumn{2}{c}{\textbf{Fine-grained grounding}} \\
\cmidrule(lr){2-4} \cmidrule(lr){5-6} \cmidrule(lr){7-8}
& RealWorldQA$^\dagger$ & MME & MMMU-Pro$^\ddagger$ & Hallusion$^\ddagger$ & POPE & HRBench8k & V$^\text{*}$ \\
\midrule
Baseline & 63.70 & 2199.5 & 31.45 & \textbf{71.50} & 87.68 & \textbf{54.13} & 66.95 \\
\rowcolor{myblue}
Confidence weighting & \textbf{66.44} & \textbf{2269.7} & \textbf{31.50} & 68.77 & 88.65 & \textbf{54.13} & \textbf{69.53} \\
Equal weighting & 64.84 & 2205.6 & 30.25 & 65.51 & \textbf{89.09} & 52.50 & 68.67 \\
\bottomrule
\end{tabular}}
\vspace{1mm}
\caption{Ablation study on learnable token weighting strategies. $^\dagger$ denotes the multiple-choice subset, and $^\ddagger$ denotes the vision split.}
\vspace{-6mm}
\label{tab:ablation_weighting}
\end{table*}

\subsubsection{MME}~\cite{fu2023mme} is a VQA benchmark in yes/no question format designed to test the hallucination of LVLMs. We follow VLMEvalKit~\cite{duan2024vlmevalkit} and append a short instruction ``please return yes or no.'' to constrain the output of VLMs, then perform string matching for the final performance score.

\subsubsection{MMMU-Pro}~\cite{yue2025mmmu} is an enhanced benchmark for VLMs designed to assess true understanding capabilities across multiple modalities. This benchmark provides VQA samples in three formats: standard textual questions with 4 options, standard textual questions with 10 options, and a purely vision-based format where the question and image are presented together in a screenshot. We evaluate our models on the vision split of the benchmark, where the entire question is posed in image format. Since the entire benchmark is MCQ-framing, we append a short instruction following VLMEvalKit to enforce the output to be the letter of the correct option, and perform string matching for the final performance.

\subsubsection{HallusionBench}~\cite{guan2024hallusionbench} is, similar to MME, a VQA benchmark in yes/no question format. This benchmark contains two parts: VQA and text QA. We focus on the VQA split, and similar to MME, a short instruction is appended at the end of each question before passing it to VLMs.

\subsubsection{POPE}~\cite{li2023evaluating} is a yes/no benchmark for testing the hallucination of VLMs. Similar to MME, we append an instruction and perform string matching for the final performance.

\subsubsection{HRBench8k~\cite{wang2025divide} and V$^\text{*}$~\cite{wu2024v}} are both VQA benchmarks featuring high-resolution image inputs. The target objects or task-relevant regions are relatively small, thus requiring strong grounding skills for VLMs to answer correctly. HRBench8k contains 800 VQA pairs and V$^\text{*}$ has around 300 samples; we include both the OCR and GPT-hard splits of V$^\text{*}$.

\section{Additional Quantitative Results}
\label{sec:quant}

\begin{figure}[t]
  \centering
  % Top Image
  \includegraphics[page=1, trim=5 5 5 5, clip, width=0.99\textwidth]{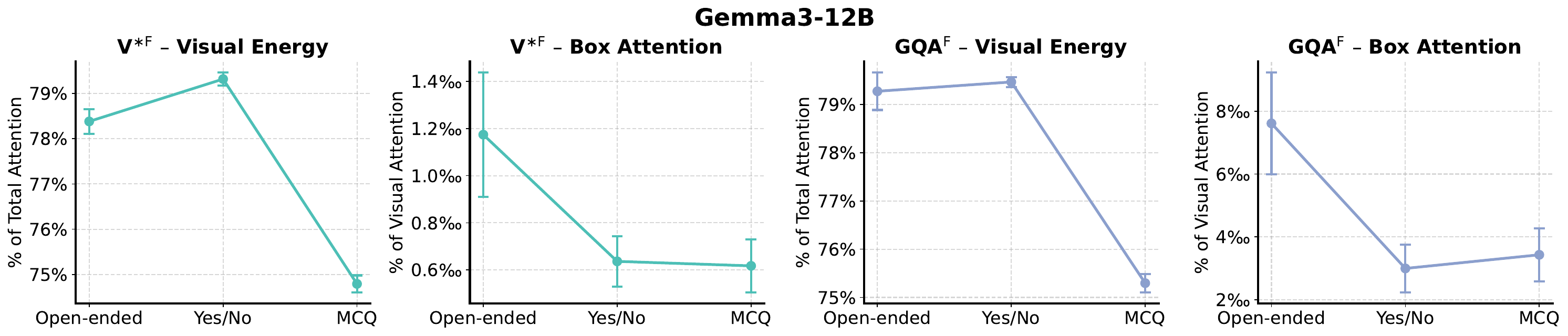}
  \\[2ex] % Adds a small vertical gap
  \includegraphics[page=1, trim=5 5 5 5, clip, width=0.99\textwidth]{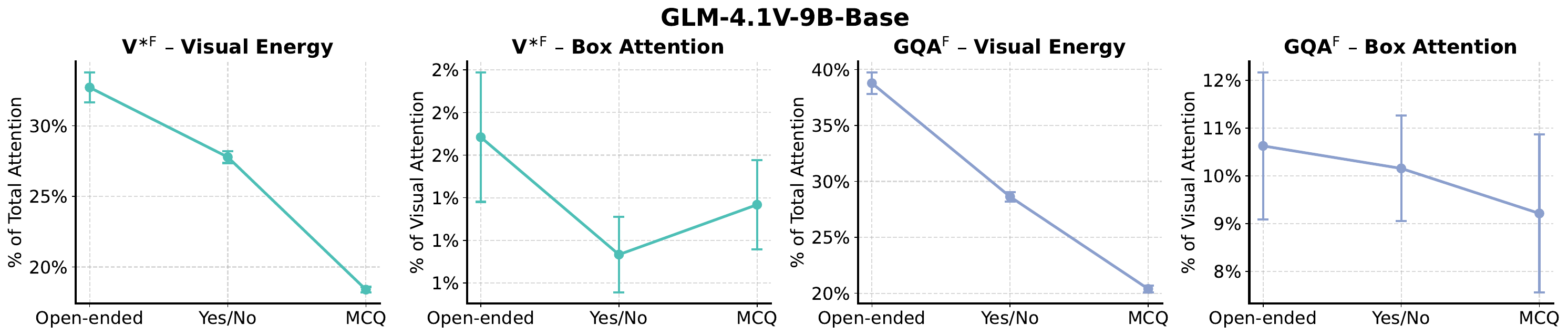}
  \\[2ex]
  \includegraphics[page=1, trim=5 5 4 5, clip, width=0.99\textwidth]{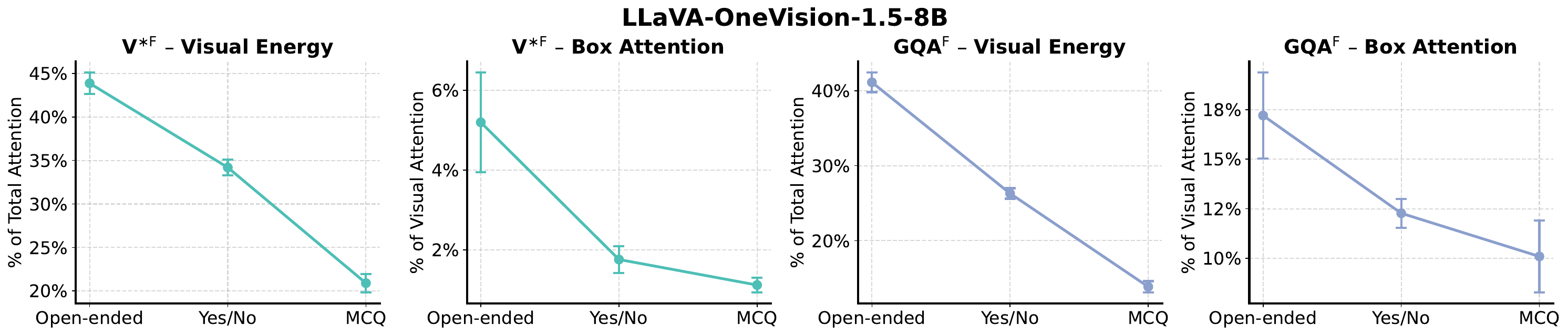}
  \\[2ex]
  % Bottom Image
  \includegraphics[page=1, trim=5 5 5 5, clip, width=0.98\textwidth]{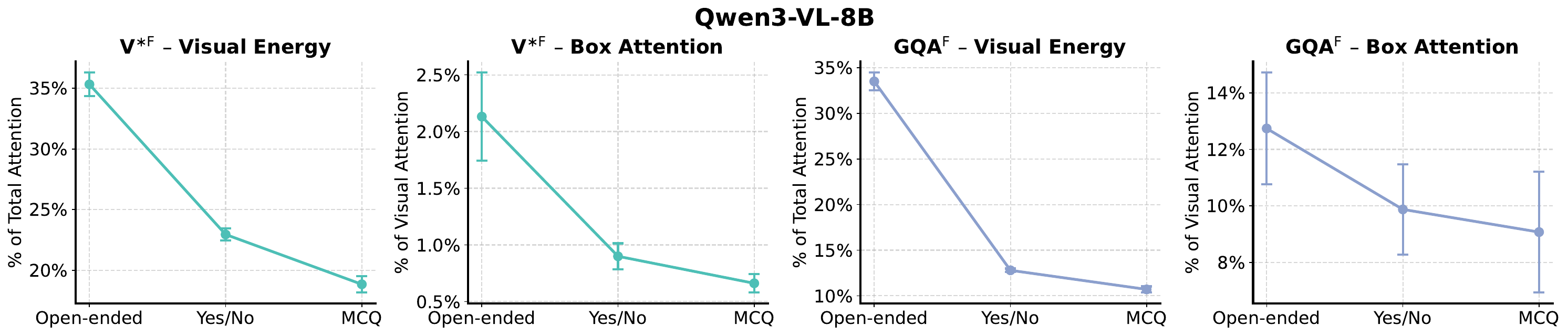}
  \vspace{-2mm}
  \caption{\textbf{Visual energy drops significantly on non-open-ended framings.}}
  \label{fig:appendix_vis_analysis}
  \vspace{-6mm}
\end{figure}

\subsection{Ablation Study}
\label{sec:ablation}

\begin{figure}[t]
  \centering
  % Top Image
  \includegraphics[page=1, trim=5 5 5 5, clip, width=0.99\textwidth]{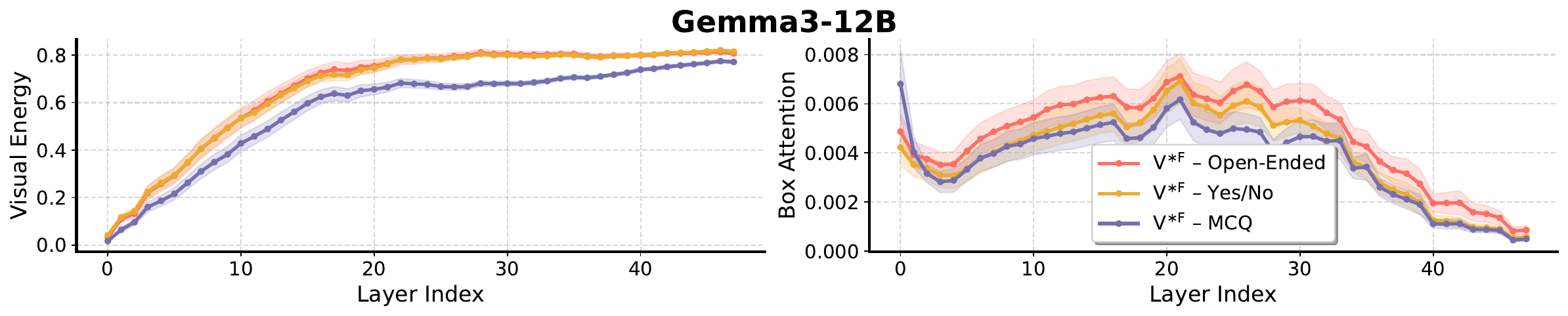}
  \\[2ex] % Adds a small vertical gap
  \includegraphics[page=1, trim=5 5 5 5, clip, width=0.99\textwidth]{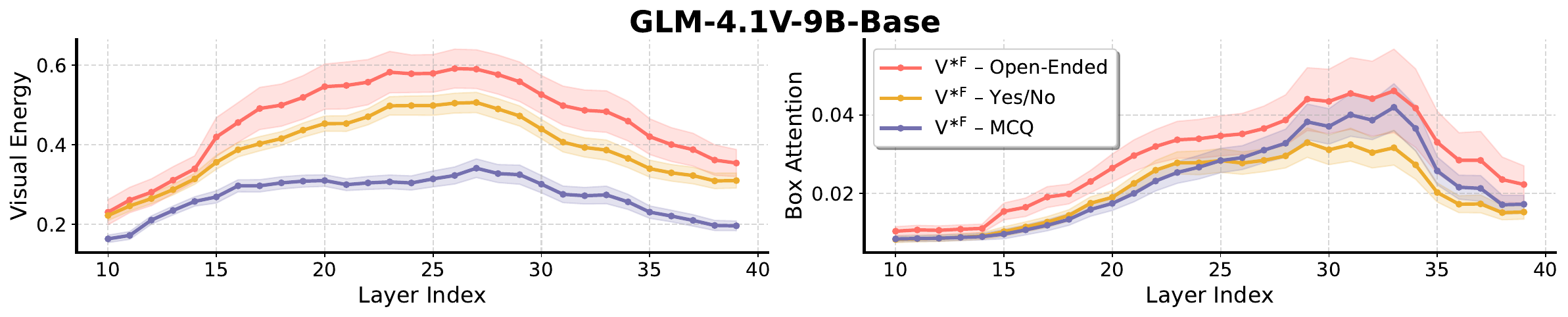}
  \\[2ex]
  \includegraphics[page=1, trim=5 5 4 5, clip, width=0.99\textwidth]{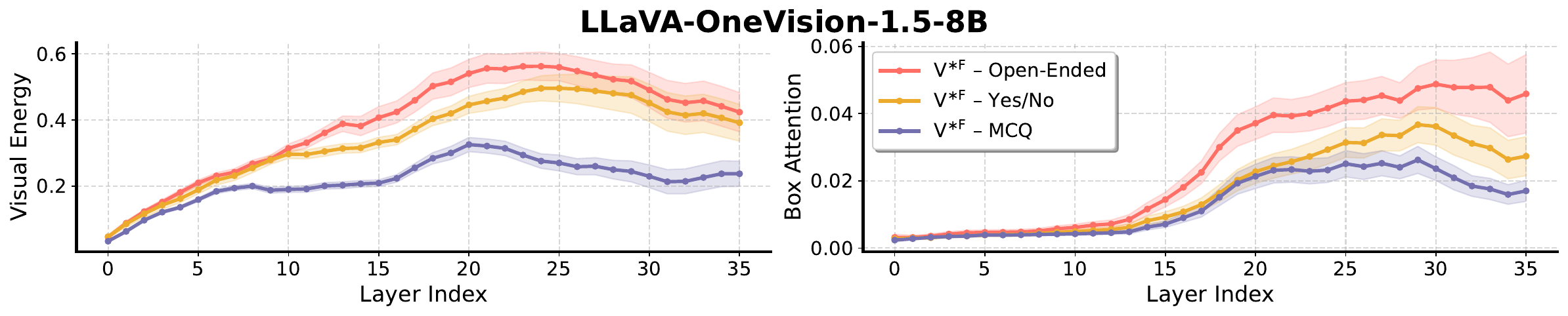}
  \\[2ex]
  % Bottom Image
  \includegraphics[page=1, trim=5 5 5 5, clip, width=0.98\textwidth]{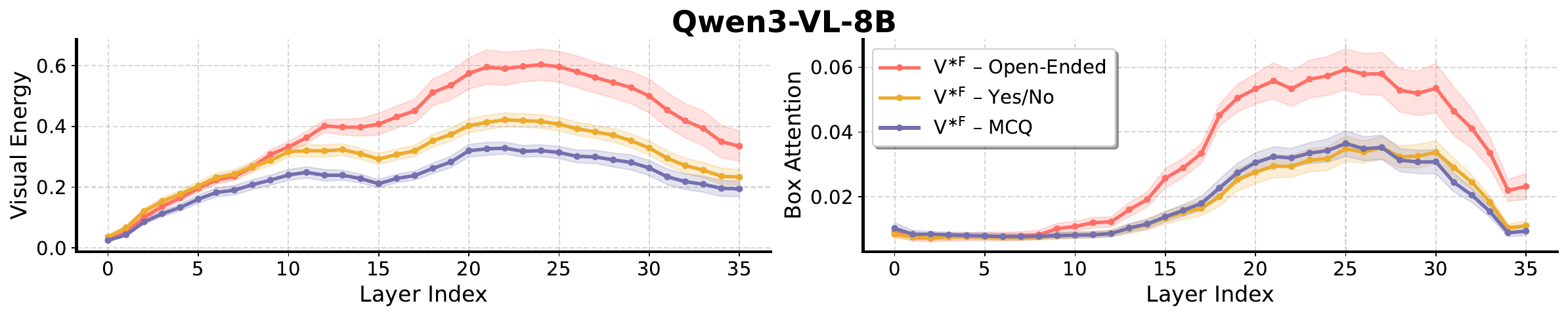}
  \vspace{-2mm}
  \caption{\textbf{Visual energy drops significantly on non-open-ended framings.}}
  \label{fig:appendix_layer_analysis}
  \vspace{-6mm}
\end{figure}

\subsubsection{Position of the Learnable Tokens}
As mentioned in~\cref{sec:methodology}, we append a set of learnable tokens to yes/no and MCQ questions to enforce VLMs to attend to the image after our attention alignment training. In this ablation, we study the impact of the position of the learnable tokens. Recall that given a VQA sample, VLMs process the input image and question into a sequence of tokens ordered as: image tokens, question tokens, and instruction tokens. Within this sequence, there are three positions to place our learnable tokens: (1) \textit{prefix}: between image and question tokens, (2) \textit{infix}: between question and instruction tokens, and (3) \textit{postfix}: after the instruction tokens at the end of the sequence. We conduct an ablation study to analyze the impact of the position, with results shown in~\cref{tab:ablation_position}. One can see that among the three positions, infix achieves the best overall performance. Our hypothesis is that since the visual attention variation is primarily driven by the question framing, placing learnable tokens immediately after the question tokens allows them to learn adjustments conditioned on the given question, while also having a more direct impact on the question compared to placing them at the end of the sequence.

\subsubsection{Weighting Strategy for Alignment Loss Across Samples}
In our prompt tuning framework, instead of applying the attention alignment loss to all training samples equally, we adopt a confidence-based weighting scheme. The motivation is that some open-ended questions may be answered incorrectly, and using attention maps from such samples would introduce inaccurate and noisy training signals. During training, we can use ground-truth labels to assess correctness and apply either a hard threshold or a soft threshold, where the latter reflects the model's confidence in its own output and thus potentially yields more reliable attention maps. To study the effectiveness of confidence weighting, we conduct an ablation comparing training with confidence weighting against training with equal weighting across all samples. As shown in~\cref{tab:ablation_weighting}, confidence weighting to filter out low-confidence samples consistently improves performance across benchmarks compared to equal weighting, verifying the effectiveness of this design choice.

\begin{figure*}[t]
  \centering
  \includegraphics[page=16, trim={10 60 10 60}, clip, width=\textwidth]{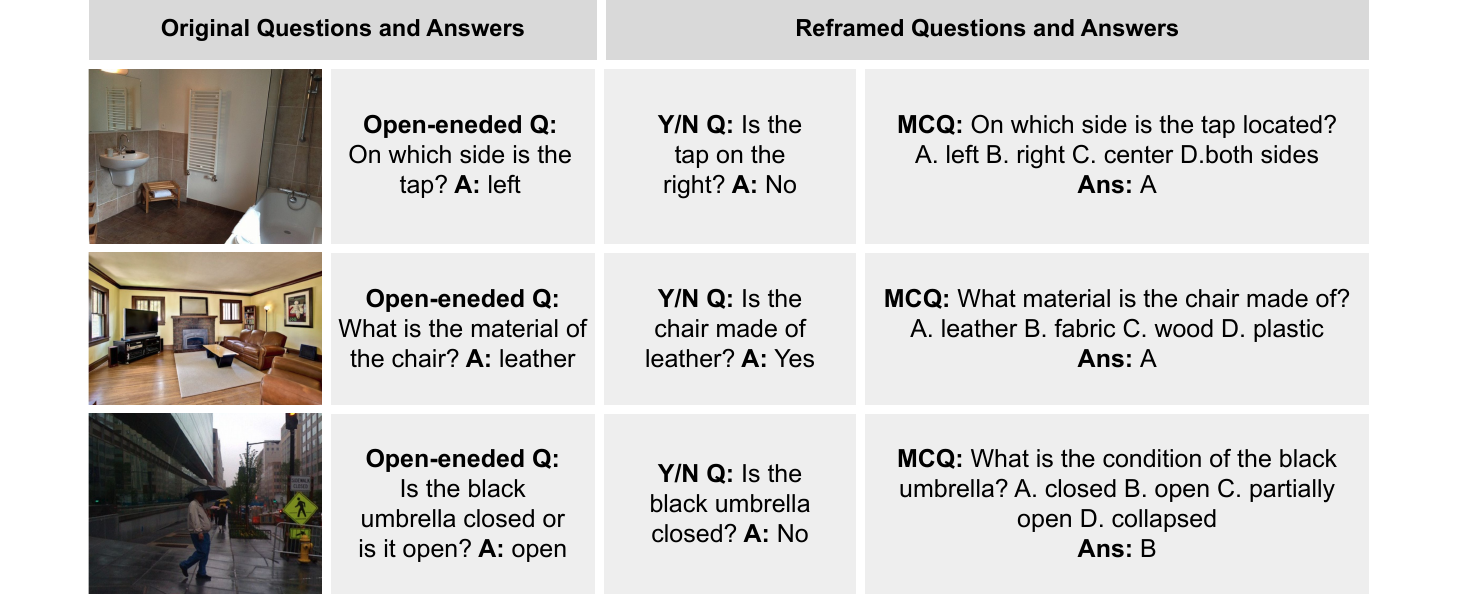}
  \vspace{-6mm}
  \caption{Qualitative examples of open-ended question reframing.}
  \label{fig:qualitative_oe}
\end{figure*}   
\begin{figure*}[t]
  \centering
  \includegraphics[page=17, trim={10 60 10 60}, clip, width=\textwidth]{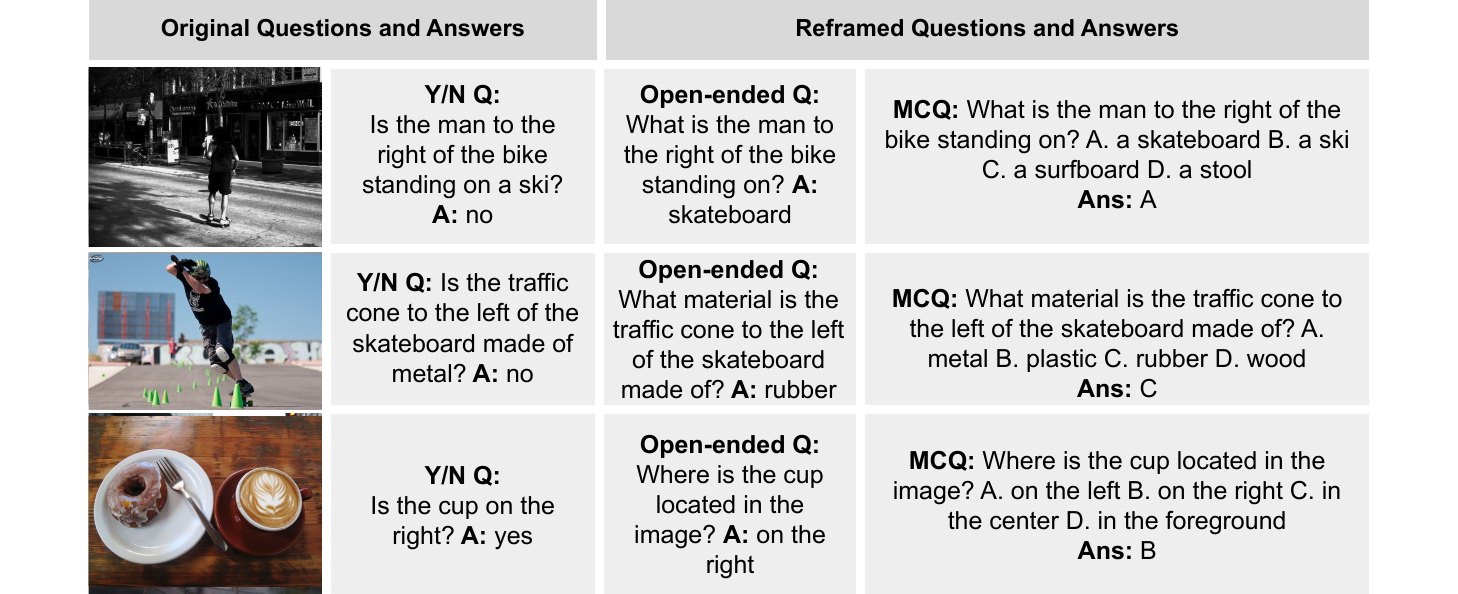}
  \vspace{-6mm}
  \caption{Qualitative examples of Yes/No question reframing.}
  \label{fig:qualitative_yn}
\end{figure*}   

\subsection{Visual Attention Analysis for Other VLMs}
\label{sec:vis_analysis}
In~\cref{sec:analysis}, we mainly focus on the behavioral study of Qwen2.5VL-7B. Here we further provide analysis on multiple models, including Gemma3~\cite{team2025gemma}, GLM-4.1V~\cite{zeng2025glm}, LLaVA-OneVision-1.5~\cite{an2025llava}, and Qwen3VL~\cite{bai2025qwen3}. The visual attention analysis and layer-wise analysis results are shown in~\cref{fig:appendix_vis_analysis} and~\cref{fig:appendix_layer_analysis}, respectively. One can see that the overall visual attention behavior across framings is similar to Qwen2.5VL-7B in the main paper on V$^\text{*F}$ and GQA$^\text{F}$, with higher bounding box attention for open-ended questions and lower attention for MCQ and yes/no questions. On Gemma3, while the bounding box attention trends are similar to our previous observation, we see that yes/no questions can potentially exhibit higher visual energy than open-ended question framings. The same behavior can be observed on GQA as well. 

\section{Dataset Examples and Qualitative Results}
\label{sec:qual}
We first provide some data samples of open-ended and Yes/No question reframing using GPT-5.1 in \cref{fig:qualitative_oe} and \cref{fig:qualitative_yn}, respectively. Data are randomly sampled from GQA$^\text{F}$; we showcase the original question in its open-ended or Yes/No framing, which is subsequently reframed into the other two formats.

\begin{figure*}[t]
  \centering
  \includegraphics[page=18, trim={10 60 10 60}, clip, width=\textwidth]{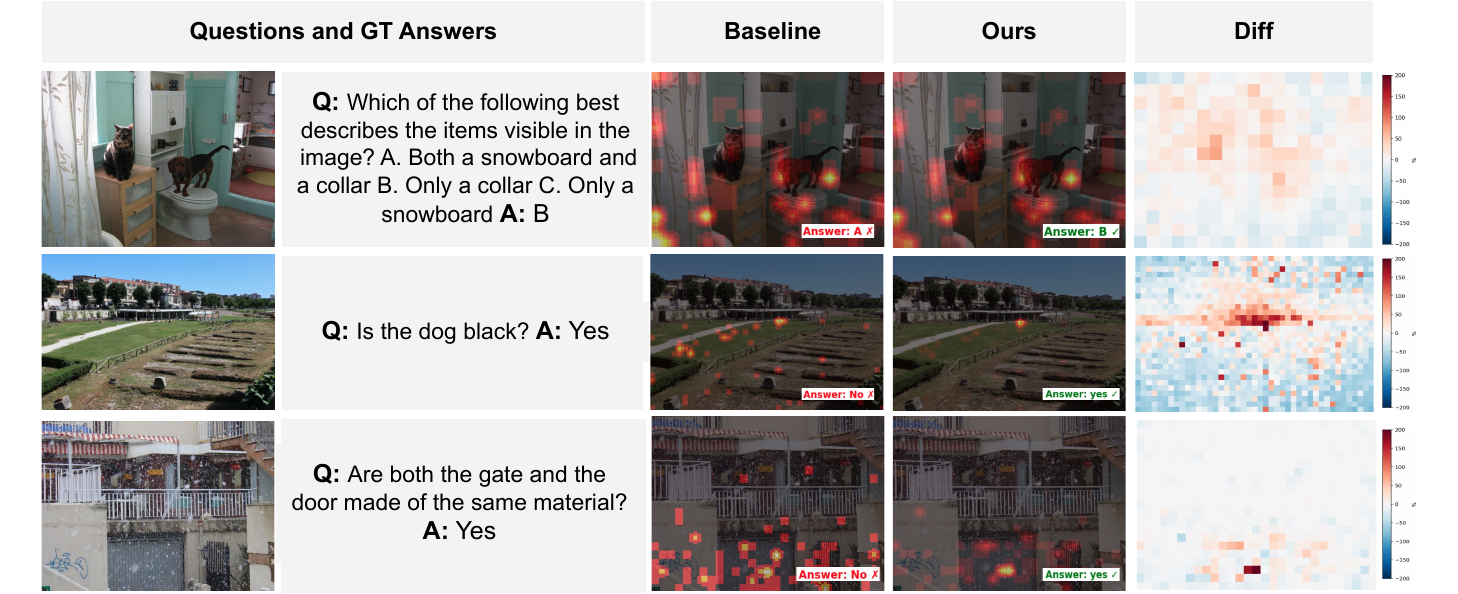}
  \vspace{-6mm}
  \caption{Qualitative comparison of Baseline and Ours (Qwen2.5-VL-7B) .}
  \label{fig:qualitative_demo}
\end{figure*}

Additionally, we provide a qualitative comparison of inference with and without our learned tokens in \cref{fig:qualitative_demo}. Given a VQA sample, we perform inference using a Qwen2.5VL-7B baseline alongside our approach with extra learned tokens, visualizing the attention rollout during output generation. As shown, the attention maps from our approach focus significantly more on task-relevant regions, whereas the baseline often spreads to irrelevant background areas. The Diff map highlights the percentage-wise change, further demonstrating that our learned tokens effectively shift attention toward the target area during inference.

\section{Limitation}
\label{sec:limitation}
While our study provides a comprehensive mechanistic analysis of framing effects across five diverse and prominent VLM families, the architectural landscape of multimodal models is rapidly evolving. Future research could extend this investigation to other emerging architectures, such as Mamba-based VLMs or Mixture-of-Experts (MoE) models, to determine if these structural paradigms inherently mitigate or exhibit similar framing-induced attention shifts.

\end{document}